\documentclass{article}

\usepackage{microtype}
\usepackage{graphicx}
\usepackage{subfigure}
\usepackage{subcaption}
\usepackage{booktabs} 

\usepackage{hyperref}

\usepackage[accepted]{icml2024/icml2024}

\usepackage{amsmath}
\usepackage{amssymb}
\usepackage{mathtools}
\usepackage{amsthm}
\usepackage{booktabs}
\usepackage{multirow}
\usepackage{siunitx} 
\usepackage{graphicx}
\usepackage{subcaption}
\usepackage{dsfont}
\usepackage{enumitem}

\usepackage{algorithm}
\usepackage{multicol} 
\usepackage{colortbl}%
\usepackage[capitalize,noabbrev]{cleveref}

\theoremstyle{plain}

\theoremstyle{definition}

\theoremstyle{remark}

\usepackage[textsize=tiny]{todonotes}

\newcommand{\ours}{\clftbo} %
\newcommand{\clftbo}{$\texttt{ifBO}$} 
\newcommand{\ftpfn}{$\texttt{FT-PFN}$} 
\newcommand{\mfpirand}{$\texttt{MFPI-random}$} 
\newcommand{\pirandh}{$\texttt{PI\,(random horizon)}$}
\newcommand{\pirandt}{$\texttt{PI\,(max, random-T)}$}
\newcommand{\mfei}{$\texttt{EI\,(one step)}$}
\newcommand{\eimax}{$\texttt{EI\,(max)}$}

\newcommand{\mfpi}{$\texttt{PI\,(one step)}$} 
\newcommand{\pimax}{$\texttt{PI\,(max)}$} 
\newcommand{\lcnet}{$\texttt{LCNet}$}
\newcommand{\dyhpo}{$\texttt{DyHPO}$}
\newcommand{\dpl}{$\texttt{DPL}$}
\newcommand{\surrmodel}{\mathcal{M}}
\newcommand{\acqpolicy}{\mathcal{A}}

\newcommand{\lcbench}{$\texttt{LCBench}$}
\newcommand{\taskset}{$\texttt{Taskset}$}
\newcommand{\pdone}{$\texttt{PD1}$}

\DeclareMathOperator*{\argmax}{arg\!max}
\DeclareMathSymbol{\shortminus}{\mathbin}{AMSa}{"39}
\DeclareMathOperator*{\loss}{f}
\DeclareMathOperator*{\bmax}{b_{\text{max}}}
\DeclareMathOperator*{\unitstep}{\mathds{1}_b}
\DeclareMathOperator*{\prior}{\pi}
\DeclareMathOperator*{\priorcurve}{\pi_{\text{curve}}}
\DeclareMathOperator*{\priorconfig}{\pi_{\text{config}}}

\newcommand{\editmode}[1]{\textcolor{black}{#1}}

\newcommand{\dotp}{
    \mathop{
        \mathchoice{\vcenter{\hbox{\LARGE$\cdot$}}}
                   {\vcenter{\hbox{\LARGE$\cdot$}}}
                   {\vcenter{\hbox{\normalsize$\cdot$}}}
                   {\vcenter{\hbox{\small$\cdot$}}}
    }
}
\newcommand{\myrowcolour}{\rowcolor[gray]{0.925}}

\icmltitlerunning{In-Context Freeze-Thaw Bayesian Optimization for Hyperparameter Optimization}

\usepackage{todonotes}
\usepackage{verbatim}

\begin{document} 

\twocolumn[
    \icmltitle{In-Context Freeze-Thaw Bayesian Optimization \\ for Hyperparameter Optimization}
    
    \icmlsetsymbol{equal}{*}
    
    \begin{icmlauthorlist}
    \icmlauthor{Herilalaina Rakotoarison}{equal,1}
    \icmlauthor{ Steven Adriaensen}{equal,1}
    \icmlauthor{Neeratyoy Mallik}{equal,1} \\
    \icmlauthor{Samir Garibov}{1}
    \icmlauthor{Edward Bergman}{1}
    \icmlauthor{Frank Hutter}{1,2}
    \end{icmlauthorlist}

    
    \icmlaffiliation{1}{Machine Learning Lab, University of Freiburg, Germany}
    \icmlaffiliation{2}{ELLIS Institute Tübingen}
    
    \icmlcorrespondingauthor{Herilalaina Rakotoarison}{rakotoah@cs.uni-freiburg.de}
    \icmlcorrespondingauthor{Steven Adriaensen}{adriaens@cs.uni-freiburg.de}
    \icmlcorrespondingauthor{Neeratyoy Mallik}{mallik@cs.uni-freiburg.de}
    
    \icmlkeywords{Machine Learning, ICML}
    
    \vskip 0.3in
]



\printAffiliationsAndNotice{\icmlEqualContribution} 

\begin{abstract}
With the increasing computational costs associated with deep learning, automated hyperparameter optimization methods, strongly relying on black-box Bayesian optimization (BO), face limitations.
Freeze-thaw BO offers a promising grey-box alternative, 
strategically allocating scarce resources incrementally to different configurations. However, the frequent surrogate model updates inherent to this approach pose challenges for existing methods, requiring retraining or fine-tuning their neural network surrogates online, introducing overhead, instability, and hyper-hyperparameters. In this work, we propose \ftpfn, a novel surrogate for Freeze-thaw style BO. \ftpfn\ is a prior-data fitted network (PFN) that leverages the transformers' in-context learning ability to efficiently and reliably do Bayesian learning curve extrapolation in a single forward pass. Our empirical analysis across three benchmark suites shows that the predictions made by \ftpfn\ are more accurate and 10-100 times faster than those of the deep Gaussian process and deep ensemble surrogates used in previous work. Furthermore, we show that, when combined with our novel acquisition mechanism (\mfpirand{}), the resulting in-context freeze-thaw BO method (\clftbo{}), yields new state-of-the-art performance in the same three families of deep learning HPO benchmarks considered in prior work.



\end{abstract}

\section{Introduction}
\label{sec:intro}

Hyperparameters are essential in deep learning (DL) to achieve strong model performance. However, due to the increasing complexity of these models, it is becoming more challenging to find promising hyperparameter settings, even with the help of hyperparameter optimization (HPO, Section~\ref{sec:hpo}) tools~\citep{feurer-automlbook19a,bischl-dmkd23a}. Traditional HPO techniques using Bayesian Optimization (BO) are unsuitable for modern DL because they treat the problem as a black box, making it computationally expensive, requiring a full model training for each evaluation. 

Recent research in HPO has shifted towards multi-fidelity methods~\cite{li-iclr17a,li-arxiv20a,falkner-icml18a,klein-arxiv20a,li-mlsys20a,awad-ijcai21a}, utilizing lower fidelity proxies (e.g., training for fewer steps, using less data, smaller models) and only evaluating the most promising hyperparameter settings at the full fidelity. While these methods have potential, they often use coarse-grained fidelity spaces and rely on the rank correlation of performances across fidelities. Moreover, they do not always fully utilize the anytime nature of algorithms such as checkpointing, continuation, and extrapolation. As a result, these methods struggle to allocate computational resources efficiently, leading to suboptimal performance in many scenarios.

The freeze-thaw BO method (Section~\ref{sec:ftbo}), originally proposed by~\citet{swersky-arxiv14a}, is a promising approach to efficiently allocate computational resources by pausing and resuming the evaluation of different hyperparameter configurations. This method is an improvement over traditional grey-box methods as it dynamically manages resources, and recent implementations~\citep{wistuba-neurips22,kadra-neurips23} hold the state-of-the-art in the low-budget regime ($\sim$ 20 full function evaluations). However, these contemporary implementations have limitations. In particular, they rely on online learning to update the surrogate model at each step which can lead high computational overhead, instability, and complexity in managing additional hyper-hyperparameters. Moreover, they also suffer from strong assumptions about learning curves, which may not be applicable in all scenarios, resulting in overly confident incorrect predictions.

\begin{figure*}[!htb]
    \centering
    \includegraphics{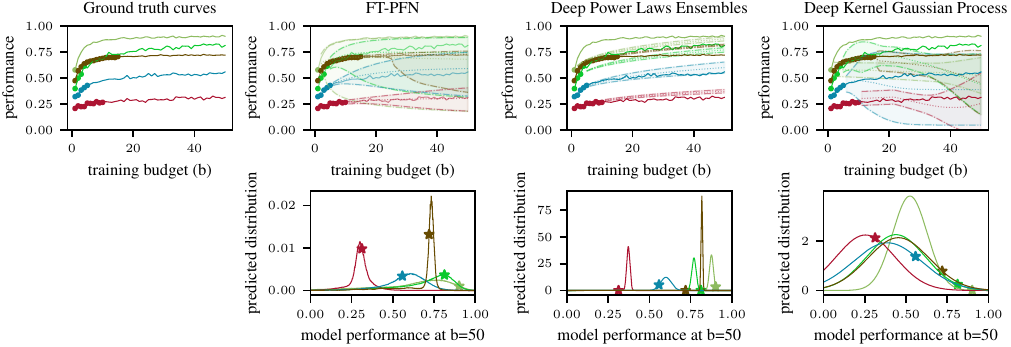}
    \caption{Comparison of freeze-thaw surrogate model predictions, given the same set of hyperparameters (HPs) and their partial learning curves. 
    The \texttt{Ground truth curves} show the real learning curves with \textit{dots} ($\dotp$) indicating the points observed as training set or context for all the surrogates.
    \clftbo{} uses \ftpfn{} as its surrogate, which requires no refitting but instead uses the \textit{training dots} as context for inferring the posterior predictive distribution of the model performance obtained at step 
    $b$ using any set of given HPs. Surrogates used in prior art, using Deep Power Laws Ensembles (\dpl{}) and Deep Kernel Gaussian Process (\dyhpo{}) respectively, are trained on the training set till convergence and then used to extrapolate the given partial curves. The bottom row shows for each surrogate, the probabilistic performance predictions made at step $50$ (last step in top row), with the \textit{stars} ({\fontfamily{lmr}\selectfont$\star$}) indicating the true value of the curve.}
    \label{fig:example}
\end{figure*}

In this work, we leverage prior-data fitted networks \citep[PFNs]{muller-iclr22a} (Section~\ref{sec:pfn}), a Transformer-based meta-learning approach to Bayesian inference, to enhance freeze-thaw BO through in-context learning. Our PFN model (\ftpfn{}) infers the task-specific relationship between hyperparameter settings and their learning curves in a single forward pass, eliminating the need for online training during the search. Figure~\ref{fig:example} compares learning curves extrapolation, including uncertainty, by our model (\ftpfn{}) and two baselines. Beyond demonstrating superior extrapolation quality, our model directly addresses key challenges in traditional HPO methods, lowering computational overhead and stabilizing the optimization process. The contributions of this paper are as follows:

\begin{itemize}
\item We propose \ftpfn{} (Section~\ref{sec:surrogate}), a new surrogate model for freeze-thaw BO, replacing online learning with in-context learning using PFNs. We train a single instance of \ftpfn{} model exclusively on synthetic data, generated from a curve prior designed to mimic realistic HPO learning curves.
\item We empirically show that \ftpfn{} outperforms existing surrogates, at point prediction and posterior distribution approximation, while being over an order of magnitude faster, and despite never having been trained on real HPO data (Section~\ref{sec:quality_surrogate}). 
\item We combine \ftpfn{} with a novel acquisition function (\mfpirand{}, Section~\ref{sec:acquisition}) and find that the resulting in-context freeze-thaw method (\clftbo{}) 
yields a new state-of-the-art performance on three benchmark suites for HPO for deep learning (\lcbench{}, \taskset{}, \pdone{}).
\end{itemize}

The code for the surrogate PFN training and reproducing experiments from this paper, is available at: \url{https://github.com/automl/ifBO}.

\section{Related Work}
\label{sec:related}

\paragraph{Multi-fidelity hyperparameter optimization} uses low-cost approximations of the objective function, for example, by evaluating only a few epochs of model training. A notable approach in this category is Hyperband~\cite{li-iclr17a}, which iteratively allocates a similar budget to various candidate hyperparameter settings and retains only the most promising ones, to be evaluated at a higher budget. Hyperband was extended for efficient parallelization~\cite{li-arxiv20a}, Bayesian optimization~\cite{falkner-icml18a} and evolutionary search~\citep{awad-ijcai21a} and to incorporate user priors~\cite{mallik2023priorband}. However, a key limitation of these multi-fidelity approaches is that they do not allow the 
continuation of 
previously discarded runs in the light of new evidence, resulting in a waste of computational resources. Additionally, their effectiveness heavily depends on a good manual choice of fidelity levels and a strong correlation between the ranks of hyperparameter settings at low and high fidelities (e.g., no crossings of learning curves).

A promising research direction to mitigate these limitations involves predicting performance at higher fidelities. \citet{domhan-ijcai15a} addressed this by proposing a Bayesian learning curve extrapolation (LCE) method. Then, \citet{klein-iclr17a} extended the latter approach to jointly model learning curves and hyperparameter values with Bayesian Neural Networks. Non-Bayesian versions of LCE have also been explored by~\citet{chandrashekaran2017speeding} and ~\citet{gargiani-automl19}. Despite the robust extrapolation capabilities of these LCE methods, fully leveraging them in guiding HPO remains a challenge.

\paragraph{Freeze-Thaw Bayesian Optimization} offers a promising solution to address the limitations of standard multi-fidelity and LCE-based HPO. Its ability to pause and resume optimization runs enables the efficient management of computational resources and ensures solid anytime performance. As a BO method, it incorporates a surrogate model to predict performance at higher fidelities and relies on the acquisition function to guide the search. 
The \emph{freeze-thaw} concept was introduced by \citet{swersky-arxiv14a}, who used a Gaussian process with exponential decay kernel for modeling learning curves and an entropy search acquisition function. \citet{wistuba-neurips22} recently proposed \dyhpo{}, an improved version that uses a learned deep kernel combined with a multi-fidelity-based Expected Improvement (EI) acquisition. Most recently, \citet{kadra-neurips23} introduced \dpl{}, another surrogate model that utilizes a deep ensemble of power laws and EI at the maximum budget as acquisition function.
These Freeze-Thaw BO methods substantially improve upon standard multi-fidelity approaches, showing particularly strong performance in low-budget settings. Nevertheless, these methods share a common challenge: The necessity for online training of a surrogate model during the search, which can be computationally intensive and may cause optimization instabilities.

\paragraph{In-context Learning (ICL)} is an exciting new learning paradigm that offers a promising alternative to online learning methods. A key 
advantage of 
ICL 
is that it does not require retraining/fine-tuning the model with new data, but instead, the data is fed to the model as a contextual prompt.
ICL 
first gained a lot of interest with the rise of Transformer-based models like large language models \citep[LLMs]{radford-openaiblog19a}
and has since also been explored 
as an end-to-end approach to black box HPO~\cite{chen-neurips22a}. 

\paragraph{Prior data fitted networks,} proposed by \citet{muller-iclr22a}, are transformer-based models that are trained to do in-context Bayesian prediction. They have been successfully used as an in-context classifier for tabular data~\citep{hollmann-iclr23a}, an in-context surrogate model for black-box HPO~\citep{muller-icml23a}, an in-context model for Bayesian learning curve extrapolation~\citep{adriaensen-neurips23}, and an in-context time-series forecaster~\citep{dooley-neurips23}. 
Our approach draws on and expands these prior works to create an efficient in-context surrogate model for freeze-thaw BO.

\section{Preliminaries}
\label{sec:preliminaries}

We now discuss some preliminaries that we build on more formally, introducing our notation along the way.
\subsection{Hyperparameter Optimization (HPO)}
\label{sec:hpo}
Consider an iterative machine learning training pipeline with configurable hyperparameters $\lambda \in \Lambda$. E.g., when training a neural network using gradient descent we could configure the learning rate, weight decay, dropout rate, etc.
Let $\loss(\lambda,b_\lambda)$ be some measure of downstream performance (e.g., validation accuracy) of the model obtained, using hyperparameter settings $\lambda$,  after $b_\lambda$ iterations of training, that we would like to maximize. HPO aims to find a hyperparameter setting producing the best model, i.e., \mbox{$
\lambda^*$: $(\lambda^*, \cdot) \in \argmax_{\lambda \in \Lambda, 1 \leq b \leq B} \loss(\lambda, b)$}, within the limited total optimization budget of $B$ iterations.
Note that in modern deep learning the budget available for HPO often does not allow us to execute more than a few full training runs. 
In this setting, the crux of HPO lies in allocating these limited training resources to the most promising hyperparameter settings, i.e., to find an allocation $\{b_\lambda\}_{\lambda \in \Lambda}$ with $b_{\lambda} \geq 0$ and $\sum_{\lambda \in \Lambda} b_{\lambda} \leq B$ that maximizes $\max_{\lambda \in \Lambda,\, 1 \leq b \leq b_{\lambda}} \loss(\lambda, b)$.

\subsection{Freeze-Thaw Bayesian Optimization}
\label{sec:ftbo}
As discussed above, 
\citet{swersky-arxiv14a} proposed to address this challenge by following a fine-grained dynamic scheduling approach, allocating resources to configurations \editmode{(and observing their performance)} ``one step at a time''. Here, one ``step'' corresponds to $\unitstep \geq 1$ iterations of model training (e.g., one epoch). 
Further, assume the maximum number of steps allocated to any single configuration $\lambda$ to be limited to $\bmax$ (i.e., training runs are limited to $\bmax \cdot \unitstep$ iterations~$\approx$ training compute per configuration). \editmode{Algorithm~\ref{algo:ftbo} shows the freeze-thaw Bayesian optimization framework that uses its history $H$, i.e., the various partial learning curves observed thus far in the current partial allocation, to fit a dynamic Bayesian surrogate model $\surrmodel{}$ that probabilistically extrapolates the partially seen performance of configuration $\lambda$ beyond 
$b_\lambda$
\editmode{(Algorithm~\ref{algo:ftbo}, line~\ref{algo:ftbo:fit})}.}
Following the BO framework, it decides which of these to continue (or start if $b_{\lambda} = 0$) using \editmode{a dynamic acquisition policy $\acqpolicy{}$} trading off exploration and exploitation \editmode{(Algorithm~\ref{algo:ftbo}, line~\ref{algo:ftbo:acq})}. 
The crucial difference \editmode{with traditional black box Bayesian Optimization} lies in that resources in the freeze-thaw framework are allocated one step, rather than one full training run, at the time. Note, this framework assumes training to be preemptive, i.e., that we can stop (\emph{freeze}) a model training, and continue (\emph{thaw}) it later; this assumption is reasonable in modern DL where checkpointing is common practice. \editmode{Also note, that black box BO can be recovered as a special case, by setting $\bmax$, the maximum number of steps allocated to any single configuration, to one, and $\unitstep$ to the maximum budget available for any single training run.}

For simplicity, in the remainder, we assume the budget step $\unitstep$ is set to 1, and the output of $\loss$ to be bounded in $[0,1]$.

\begin{algorithm}[tbp]
 
\small
  \caption{Freeze-thaw Bayesian Optimization. \textcolor{blue}{Blue comments detail \clftbo\ specifics.}\label{algo:ftbo}}
  \textbf{Input:} $\Lambda$\textcolor{gray}{: configuration space,} \\
  \hspace*{0.95cm}$\loss$\textcolor{gray}{: measure of model performance to be maximized,} \\
  \hspace*{0.95cm}$\mathds{1}_b$\textcolor{gray}{: iterations of model training per freeze-thaw step,}\\
    \hspace*{0.95cm}$\bmax$\textcolor{gray}{: maximal steps  for any configuration $\lambda \in \Lambda$,}\\
    \hspace*{0.94cm}$B$\textcolor{gray}{: total HPO budget in iterations of model training.}\\ 
  \textbf{Components:} \\
  \hspace*{0.3cm}$\surrmodel{}$\textcolor{gray}{: the dynamic surrogate model (\textcolor{blue}{\ftpfn{}}),} \\
  \hspace*{0.3cm}$\acqpolicy{}$\textcolor{gray}{: the dynamic acquisition policy (\textcolor{blue}{\mfpirand{}, Alg.~\ref{algo:mf-af}})} \\
    \textbf{Output:} $\lambda^*$\textcolor{gray}{$ \in \Lambda$, obtaining the best observed performance} \\
    
  \textbf{Procedure:} $\text{HPO}$($\Lambda$, $\loss$, $\unitstep$, $\bmax$, $B$): \\
  \begin{algorithmic}[1]
      \STATE $b_{\lambda'} \leftarrow 0$, $\forall \lambda'\in\Lambda$
      \STATE $\lambda \sim \mathcal{U}(\Lambda) \hfill$~{\color{gray}initial random sample}
      \STATE $b_\lambda \leftarrow \unitstep$
      \STATE $H \leftarrow{} \{(\lambda, b_\lambda, \loss(\lambda,b_\lambda)\}\hfill$~{\color{gray}evaluate $\loss$ (train $\lambda$ for first step)}

      \item[]
      
      \WHILE{$|H| \cdot \mathds{1}_b < B$}\label{algo:ftbo:startloop}
      
        \STATE {Train $\surrmodel{}$ on $H$}\hfill{\color{blue} \ftpfn\ requires no model fitting}\label{algo:ftbo:fit}

        \STATE $\lambda \xleftarrow{}$ $\acqpolicy{}$($\Lambda,\surrmodel{},H,\bmax$)\hfill~{\color{gray}select $\lambda$ to thaw}\label{algo:ftbo:acq}
        \STATE $b_\lambda \leftarrow b_\lambda + \mathds{1}_b$\hfill{\color{gray}thaw $\lambda$ for one step}
        \STATE $y \leftarrow{} \loss(\lambda, b_\lambda)$\hfill{\color{gray}evaluate $\loss$}

        \STATE $H \leftarrow H \cup \{(\lambda, b_\lambda, y)\}$

    \ENDWHILE \label{algo:ftbo:endloop}

    \item[]
    

    \STATE \textbf{return} \editmode{ $\lambda^*$: $(\lambda^*,\cdot) \in\argmax_{\lambda \in \Lambda,\, 1 \leq b \leq b_{\lambda}} \loss(\lambda, b )$}
\end{algorithmic}
\end{algorithm}

\subsection{Prior-data Fitted Networks (PFNs)}
\label{sec:pfn}
As briefly discussed above, PFNs~\citep{muller-iclr22a} are neural networks $q_\theta$ that are trained to do Bayesian prediction for supervised learning in a single forward pass. 
More specifically, let $D = D_\text{train} \cup \{(x_\text{test},y_\text{test})\}$ be a dataset used for training; the PFN's parameters $\theta$ are optimized to take $D_\text{train}$ and $x_\text{test}$ as inputs and make predictions that approximate
the posterior predictive distribution (PPD) of the output label $y_\text{test}$: \[q_\theta(x_\text{test},D_\text{train}) \approx \mathbb{P}(y_\text{test}\,|\,x_\text{test},D_\text{train}),\] 
in expectation over datasets $D$ sampled from a prior $p(\mathcal{D})$ over datasets.
At test time, the PFN does not update its parameters given a training dataset $D_\text{train}$, but rather takes $D_\text{train}$ as a contextual input, predicting the labels of unseen examples through \emph{in-context learning}. The PFN is pretrained once for a specific prior $p(\mathcal{D})$ and used in downstream Bayesian prediction tasks without further fine-tuning. More specifically, it is trained to minimize the cross-entropy for predicting the hold-out example\textquotesingle{s} label $y_\text{test}$, given $x_\text{test}$ and $D_\text{train}$:
\begin{align}
\label{eq:celoss}
\begin{split}
\ell_{\theta} = \mathbb{E}[ \shortminus \text{log} \  q_{ \theta}(y_\text{test} |x_\text{test}, D_\text{train}) ] \\
\text{with \quad} \{(x_\text{test}, y_\text{test})\} \cup D_\text{train} \sim p(\mathcal{D})
\end{split}
\end{align}
\citet{muller-iclr22a} proved that this training procedure coincides with minimizing the KL divergence between the PFN\textquotesingle{s} predictions and the true PPD. 

\section{In-Context Freeze-Thaw BO (\ours{})}
In this section, we describe \ours{}, the in-context learning variant of the freeze-thaw framework that we propose as an alternative for the existing online learning implementations~\cite{wistuba-neurips22,kadra-neurips23}. As can be seen in Algorithm~\ref{algo:ftbo} (line~\ref{algo:ftbo:fit}), the critical difference lies in the fact that we do not need to refit our surrogate model after every allocation step. 
By skipping the online refitting stage, we reduce computational overhead, code complexity, and hyper-hyperparameters. Note that within the freeze-thaw BO framework described in Section~\ref{sec:ftbo}, our method is fully characterized by our choice of surrogate model (Section~\ref{sec:surrogate}) and dynamic acquisition \editmode{policy} (Section~\ref{sec:acquisition}).

\subsection{Dynamic Surrogate Model (\ftpfn{})}
\label{sec:surrogate}
We propose \ftpfn{} a prior-data fitted network \citep[PFNs]{muller-iclr22a} trained to be used as an in-context dynamic surrogate model in the freeze-thaw framework. As described in Section~\ref{sec:pfn}, PFNs represent a general meta-learned approach to Bayesian prediction, characterized by the data used for meta-training. 
Following previous works using PFNs~\citep{muller-iclr22a,hollmann-iclr23a,muller-icml23a,adriaensen-neurips23,dooley-neurips23}, we train the PFN only on \emph{synthetically generated data}, allowing us to generate virtually unlimited data and giving us full control over any biases therein.
We aim to train \ftpfn{} on artificial data ($D$) that resembles the real performance data we observe in the context of freeze-thaw Bayesian optimization, i.e., a collection of learning curves of training runs for the same task, but using different hyperparameter settings, i.e., $$\bigcup_{\lambda\in\Lambda}
\biggl\{\bigl((\lambda,1), \loss(\lambda,1)\bigl),\ldots,\bigl((\lambda,\bmax), \loss(\lambda,\bmax)\bigl)\biggl\}$$




\paragraph{Prior desiderata:} From a Bayesian perspective, we want to generate data from a prior data model that captures our beliefs on the relationship between hyperparameters $\lambda$, training budget $b$, and model performance $\loss(\lambda,b)$. While one could design such prior for a specific HPO scenario, our goal here is to construct a generic prior, resulting in an \ftpfn{} surrogate for general HPO. To this end, we leverage general beliefs, e.g., we expect learning curves to be noisy, but to exhibit an improving, convex, and converging trend; curves on the same task are expected to have similar start, saturation, and convergence points; and training runs using similar hyperparameter settings to produce similar learning curves. 


\begin{figure}
  \includegraphics[height=4cm]{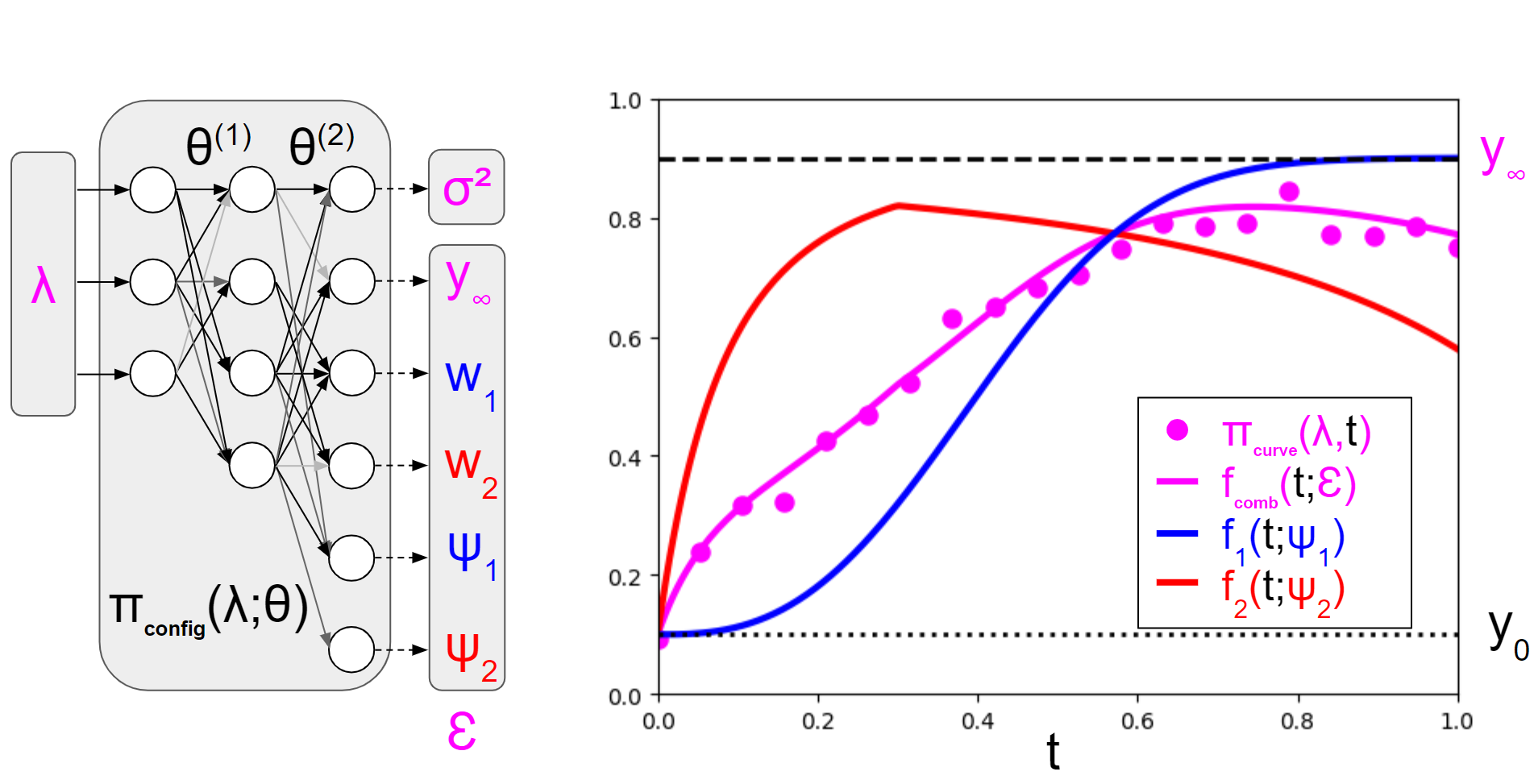}
  \caption{Diagram for the prior data model described in Section~\ref{sec:surrogate} that was used to generate data for meta-training \ftpfn{}. On the left, we have the randomly initialized neural network $\priorconfig$ that models the relationship between a hyperparameter setting $\lambda$ and its learning curve (shown in pink), whose output parameterizes a curve model $\priorcurve$ that is a linear combination of $K$ (=2 in this illustration) basis functions (shown in red and blue) with added $\lambda$-specific Gaussian noise with variance $\sigma^2$.\label{fig:prior_model}}
\end{figure}

\paragraph{Prior data model:} Following \citet{klein-iclr17a} and \citet{kadra-neurips23}, we model the performance curve of a hyperparameter $\lambda$ using a parametric curve model $\prior_{\text{curve}}$, whose parameters are sampled from an another prior model $\prior_{\text{config}}$ taking the hyperparameter $\lambda$ as input (see Figure~\ref{fig:prior_model}). Following \citet{domhan-ijcai15a} and \citet{adriaensen-neurips23}, we define $\priorcurve$ as a weighted combination of $K$ basis functions $f_k$, with additive Gaussian noise. Thus, the parameters of $\priorcurve$ include $\mathcal{E}$ (the set of parameters and weight of all basis functions), along with $\sigma^2$ (noise). As for $\priorconfig$, we adopt a neural network with weights $\theta$. Unlike previous works, we do not train the weights $\theta$ of this neural network. Instead, we randomly initialize the network, to represent a task-specific relationship between hyperparameters and their learning curves, which we then use to generate data for training \ftpfn{}. This can be viewed as generating samples from a Bayesian Neural Network (BNN) prior, meta-training \ftpfn{} to emulate \lcnet{}-like~\citep{klein-iclr17a} BNN inference through in-context learning. 

Formally, we define the performance of a hyperparameter $\lambda$ at a training time $t$ as follows:

\begin{align}
&\priorcurve(\lambda, t) \sim \mathcal{N}\left( f_{\text{comb}}(t; \mathcal{E}), \sigma^2 \right) \label{eq:priorcurve} \\
&\text{with} \quad f_{\text{comb}}(t; \mathcal{E}) = y_0 + (y_{\infty} - y_0) \cdot \sum_{k=1}^K w_k f_k(t; \Psi_k) \notag \\
&\text{and} \quad (\sigma^2, \underbrace{\left( y_{\infty}, w_1, \ldots, w_K, \Psi_1, \ldots, \Psi_K \right)}_{\mathcal{E}}) \sim \priorconfig(\lambda; \theta) \notag,
\end{align}

where $y_0$ is the initial model performance\footnote{Note that $y_0 \notin \mathcal{E}$ as we assume it to be independent of $\lambda$.}, and $y_{\infty}$ that at convergence. $w_k$ is the weight of basis curve $f_k$, and $\Psi_k$ its basis function specific parameters. In this work, we adopt four different basis functions ($K=4$), each having four parameters, resulting in a total of 22 ($=|\mathcal{E}|+1$) parameters depending on $\lambda$ through $\priorconfig$. Our four basis functions subsume the power law model used by~\citet{kadra-neurips23}, all three basis functions used by~\citet{adriaensen-neurips23}, and 9 of the 11 basis functions originally proposed by~\citet{domhan-ijcai15a}.\footnote{Our model excludes the two unbounded basis curves.} Furthermore, unlike those considered in previous works, our basis functions can have a breaking point~\citep{caballero-iclr23} at which convergence stagnates or performance diverges, resulting in a more heterogeneous and realistic model.

To sample a collection of curves on the same task from our prior data model, we (i) sample their hyperparameters from a unit hypercube; (ii) initialize $\priorconfig$ as in \cite{muller-icml23a}; then (iii) apply Equation~\ref{eq:priorcurve} to obtain the performance of each hyperparameter $\lambda$ at a given training time $t$. To reduce the entropy of this prior (and PFN training time), we assume the training time $t$ to be normalized in $[0,1]$. In practice, we also sample a maximum training time $\bmax$ per task and define $t_b=\frac{b}{\bmax}$ for $b\in[1,\bmax]$. Further details about meta-training \ftpfn{} can be found in Appendix~\ref{a:ftpfn}, including the basis curves, their parameters, and illustrations of samples from our learning curve prior (Appendix~\ref{a:lcprior}); a detailed description of the procedure we use for generating our meta-training data (Appendix~\ref{a:priorgen}); and the architecture and hyperparameters used (Appendix~\ref{a:pfntraining}).

\subsection{Dynamic Acquisition Policy (\mfpirand{})}
\label{sec:acquisition}

Following the Freeze-Thaw Bayesian Optimization framework (Section~\ref{sec:ftbo}), we continue training the configuration that maximizes some acquisition function (AF), i.e., $\argmax_{\lambda\in\Lambda} \texttt{AF}(\lambda)$. Conceptually, we would like to continue the training that is \emph{most likely} to \emph{quickly} produce a model performing \emph{substantially better} than the best model obtained thus far. This notion can be formalized as an acquisition function,
\begin{equation}\label{eq:mfpi}
    \texttt{MFPI}(\lambda; h, T) = \mathds{P}(\surrmodel{}(\lambda,~\min(b_{\lambda} + h, \bmax)) > T)
\end{equation}
which evaluates to the predicted likelihood that a candidate configuration $\lambda \in \Lambda$ after $h$ \emph{more} steps of training obtains a model exceeding the $T$ performance threshold. 
Here, $\surrmodel{}$ is the trained surrogate model, and the extrapolation horizon $1 \leq h \leq \bmax$ is a parameter controlling the trade-off between immediate and long-term gains, i.e. \emph{what is quick enough}, and the threshold $f_{\text{best}} \leq T < 1$ a parameter controlling the trade-off between high and low risk/gain, i.e., \emph{what improvement is substantial}.
Note that for $h = \bmax$ and $T = f_{\text{best}}$, we recover the Probability of Improvement (PI) acquisition function~\citep{mockus-tgo78a}.
The values we choose for these hyper-hyperparameters will affect which configuration gets continued 
(see Figure~\ref{fig:mfpi_example}, Appendix~\ref{app:af}). Their optimal settings depend on the desired freeze-thaw behavior and are not straightforward to determine. It might even be beneficial to adjust them dynamically during the run. Instead of using a fixed performance threshold $T$ or a fixed extrapolation horizon $h$~\citep{wistuba-neurips22,kadra-neurips23}, we explore a range of possible thresholds and horizons by randomizing these.
Such random sampling procedure is undertaken every freeze-thaw BO iteration and is akin to an AF selection from a portfolio of different $\texttt{MFPIs}$. 
We posit that this hedging with a portfolio of AFs
(portfolio of Equation~\ref{eq:mfpi} with different $h,~T$) in each iteration benefits freeze-thaw setups where queries are more granular than standard BO.  
The result is a simple, parameter-free AF with a balanced exploration-exploitation trade-off,
\begin{equation} 
\begin{split}
\label{eq:mfpi_random}
    \texttt{MFPI-random}(\lambda) = \texttt{MFPI}(\lambda; h^{\text{rand}},~T^{\text{rand}}) \\
    \text{with }
    h^{\text{rand}} \sim \mathcal{U}(1, \bmax) \text{ and } \\
    T^{\text{rand}} = f_{\text{best}} + \tau^{\text{rand}} \cdot (1 - f_{\text{best}}) \\
    \text{ with } log_{10}(\tau^{\text{rand}}) \sim \mathcal{U}(-4, -1)
\end{split}
\end{equation}
Further details, as well as pseudo code (Algorithm~\ref{algo:mf-af}) can be found in Appendix~\ref{app:af}.

\newsavebox\CBox
\def\mathBF#1{\sbox\CBox{#1}\resizebox{\wd\CBox}{\ht\CBox}{\ensuremath{\mathbf{#1}}}}
\newcommand{\best}[1]{\mathBF{#1}}

\begin{table*}[htbp]
\centering
\caption{
Comparison of \ftpfn, a variant of \ftpfn\ that excludes hyperparameters, \dyhpo\ and \dpl\ across three benchmarks. Values represent the median over tasks of the log-likelihood and mean squared error (MSE) as well as the runtime of predictions. }\label{tab:quality_predictions}  
\resizebox{\textwidth}{!}{\begin{tabular}{
  l
  l
  S[table-format=-2.3]
  S[table-format=1.3]
  S[table-format=-2.3]
  S[table-format=1.3]
  S[table-format=3.3, table-comparator=true]
  S[table-format=1.3]
  S[table-format=2.3]
}
\toprule
& & \multicolumn{2}{c}{LCBench} & \multicolumn{2}{c}{PD1} & \multicolumn{2}{c}{Taskset} & \multicolumn{1}{c}{\textbf{Runtime (s)}} \\
\cmidrule(lr){3-4} \cmidrule(lr){5-6} \cmidrule(lr){7-8} 
\textbf{\# samples} & \textbf{Method} & {\textbf{Log-likelihood}} & {\textbf{MSE}} & {\textbf{Log-likelihood}} & {\textbf{MSE}} & {\textbf{Log-likelihood}} & {\textbf{MSE}} & {\textbf{}} \\
\midrule

 & DPL & -14.577 & 0.007 & -13.384 & 0.043 & -26.011 & 0.005 & 17.686 \\
400 & DyHPO & -0.481 & 0.042 & -0.573 & 0.104 & -0.465 & 0.009 & 16.860 \\
 & FT-PFN (no HPs) & 1.649 & 0.008 & 0.983 & 0.028 & 2.860 & 0.005 & 0.215 \\
 & \textbf{FT-PFN} & 1.876 & 0.005 & 0.925 & 0.030 & 2.934 & 0.004 & 0.225 \\

\myrowcolour
 & DPL & -13.291 & 0.007 & -11.721 & 0.037 & -21.779 & 0.005 & 33.480 \\
\myrowcolour
800 & DyHPO & -0.426 & 0.031 & -0.510 & 0.088 & -0.419 & 0.008 & 64.809 \\
\myrowcolour
 & FT-PFN (no HPs) & 1.701 & 0.007 & 1.103 & 0.024 & 2.835 & 0.005 & 0.527 \\
\myrowcolour
 & \textbf{FT-PFN} & 2.044 & 0.004 & 1.072 & 0.025 & 2.975 & 0.004 & 0.541 \\
 
 & DPL & -11.983 & 0.007 & -11.017 & 0.035 & -20.350 & 0.004 & 41.956 \\
1000 & DyHPO & -0.368 & 0.012 & -0.457 & 0.071 & -0.381 & 0.008 & 59.949 \\
& FT-PFN (no HPs) & 1.763 & 0.007 & 1.120 & 0.024 & 2.877 & 0.005 & 0.687 \\
 & \textbf{FT-PFN} & 2.118 & 0.004 & 1.133 & 0.024 & 3.016 & 0.004 & 0.719 \\

\myrowcolour
 & DPL & -11.333 & 0.007 & -10.353 & 0.033 & -17.760 & 0.004 & 56.576 \\
 \myrowcolour
1400 & DyHPO & -0.361 & 0.011 & -0.438 & 0.061 & 
-0.374 & 0.008 & 112.168 \\
\myrowcolour
 & FT-PFN (no HPs) & 1.733 & 0.007 & 1.225 & 0.021 & 2.874 & 0.005 & 1.084 \\
\myrowcolour
 & \textbf{FT-PFN} & 2.137 & 0.003 & 1.201 & 0.022 & 3.042 & 0.004 & 1.130 \\

 & DPL & -9.182 & 0.007 & -9.263 & 0.035 & -13.712 & 0.004 & 73.435 \\
1800 & DyHPO & -0.365 & 0.009 & -0.437 & 0.058 & -0.381 & 0.008 & 166.491 \\
& FT-PFN (no HPs) & 1.753 & 0.006 & 1.251 & 0.019 & 2.858 & 0.005 & 1.635 \\
 & \textbf{FT-PFN} & 2.199 & 0.003 & 1.192 & 0.022 & 3.057 & 0.004 & 1.715 \\

\bottomrule
\end{tabular}}
\end{table*}

\section{Experiments}
\label{sec:exp}

In this section, we compare \ours\ to state-of-the-art multi-fidelity 
freeze-thaw Bayesian optimization methods.
To this end, we first assess \ftpfn{} in terms of the quality and cost of its prediction (Section~\ref{sec:quality_surrogate}). Then, we assess our approach, which combines \ftpfn\ with our \mfpirand\ acquisition function, on HPO tasks (Section~\ref{sec:sota}). Finally, we conduct an ablation study on acquisition function used in \ours\ (Section~\ref{sec:ablate_acquisition}).

We conduct our experiments on three benchmarks: \lcbench\ \cite{zimmer-tpami21a}, \pdone{}~\cite{wang-arxiv21a}, and \taskset~\cite{metz-arxiv20b}. These benchmarks, covering different architectures (Transformers, CNNs, MLPs) and tasks (NLP, vision, tabular data), are commonly used in the HPO literature.
A detailed overview of the tasks included in each benchmark is presented in Appendix~\ref{a:benchmarks}.

Our main baselines are both other recent freeze-thaw approaches: \dyhpo\ \cite{wistuba-neurips22} and \dpl\ \cite{kadra-neurips23}. 
We reimplement the above two baselines in order to allow ablation of the online learning surrogates with different acquisition functions. 
Refer to Appendix~\ref{app:baselines} for more details.

\subsection{Cost and Quality of Predictions}
\label{sec:quality_surrogate}

In this section, we compare the predictive capabilities of \ftpfn\ to that of existing surrogate models, including the deep Gaussian process of \dyhpo{}~\cite{wistuba-neurips22} and the deep ensemble of power laws model of \dpl{}~\cite{kadra-neurips23}. We also consider a variant of \ftpfn\ trained on the same prior, but not taking the hyperparameters as input (referred to as ``no HPs"). This variant bases its predictions solely on a set of partially observed learning curves.

\paragraph{Evaluation procedure:} From a given benchmark, we sample both a set of partial curves, where each curve has its own set of target epochs. The selection process is strategically designed to encompass a wide range of scenarios, varying from depth-first approaches, which involve a smaller number of long curves, to breadth-first approaches, where multiple shorter curves are explored. Additional details on the sampling strategy can be found in Appendix~\ref{a:priorgen}. 
To assess the quality of the predictions, we utilize two metrics: log-likelihood (log-score, the higher the better), measuring the approximation of the posterior distribution ($\sim$ uncertainty calibration), and mean squared error (the lower the better), measuring the accuracy of point predictions. We also report the runtime, accounting for fitting and inference of each surrogate. The evaluation was run on a single \texttt{Intel Xeon 6242 CPU}.

\paragraph{Results discussion:} Table~\ref{tab:quality_predictions} presents the log-likelihood and MSE (Mean Squared Error) for each approach relative to the context sample size. As expected, we observe an increase in log-likelihood and a decrease in MSE as the context size get larger. Notably, \ftpfn\ and its No HPs variant significantly outperform \dpl\ and \dyhpo\ in terms of log-likelihood. \dpl\ in particular has low log-likelihood values, corresponding to a poor uncertainty estimate such as being overly confident in incorrect predictions. This may be due to the very low ensemble size ($=5$) adopted by \cite{kadra-neurips23} compounded by their strong power law assumption.
On the other hand, \dyhpo\ struggles with low log-likelihood due to its inability to extrapolate beyond a single step effectively. 
Regarding MSE, \ftpfn\ generally surpasses the baselines in \lcbench{} and \pdone{}, performing comparably to \dpl\ on \taskset{}. 

\begin{figure*}[ht]
      \includegraphics[width=\textwidth]{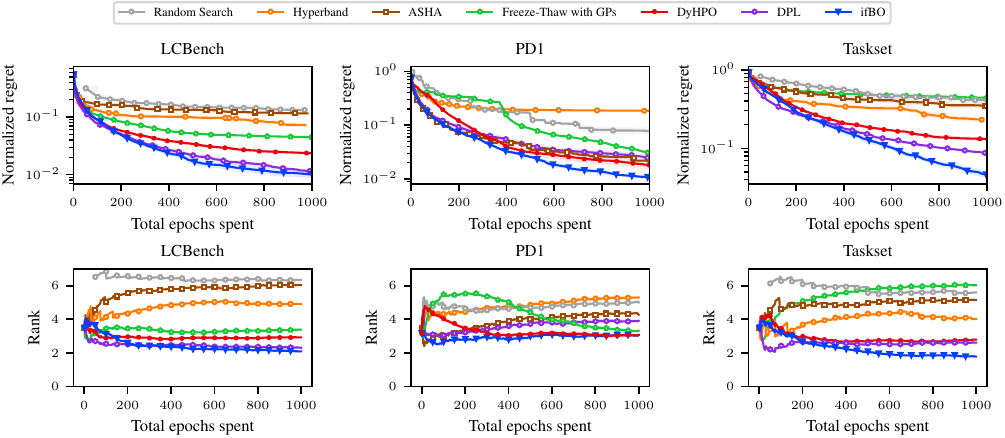}\centering
  \caption{Comparison of our method against state-of-the-art baselines on all 3 benchmarks. First row shows normalized regret aggregated across multiple tasks in each benchmark (See Appendix \ref{a:benchmarks} for benchmark details, and the results per task can be found in Appendix~\ref{a:per_task_plot}). Second row shows the average ranks of each method. \label{fig:per-bench-sota}}
\end{figure*}

Beyond the impressive log-likelihood and MSE results, our approach also yield significant speed advantages over the baseline methods. \editmode{Importantly, \ftpfn\ maintains superiority in quality and speed for inferences with more than $1000$ samples as a context without not being trained in this regime.} Depending on the context sample size, our method achieves speedups ranging from $10\times$ to $100\times$ faster than \dpl\ and \dyhpo.  

\subsection{Assessment of HPO performance}
\label{sec:sota}

In this section, we present an extensive empirical comparison, showing the merits of our method on HPO tasks across a variety of \textbf{tabular} benchmarks (see, Appendix~\ref{a:benchmarks}) in the low budget regime. Additionally to \dpl\ and \dyhpo, we include Hyperband, ASHA, Gaussian process-based FT-BO (using \dyhpo's one-step EI acquisition function) and uniform random search, as baselines (see, Appendix~\ref{app:baselines}). Each algorithm is allocated a total budget of 1000 steps for every task, which corresponds to 20 full trainings on \lcbench\ and \taskset\, with each task being repeated 10 times~with different seeds, each time starting from a different random configuration (see Algorithm~\ref{algo:ftbo}). 
For \pdone\, tasks have varying maximum steps allowed ($\bmax$), but for consistency and fair aggregation across benchmarks, we applied the same HPO budget of 1000 here too.
We report two complementary metrics: The normalized regret, capturing performance differences, and the average rank of each method, capturing the relative order. Formally, the normalized regret corresponds to a $[0, 1]$ normalization of the observed error w.r.t to the best (lowest) and worst (highest) errors recorded by all algorithms on the task. 

\paragraph{Results discussion:}  Figure~\ref{fig:per-bench-sota} presents the comparative results per benchmark family.  The results validate the superiority of freeze-thaw approaches (\ours, \dyhpo, and \dpl) compared to standard approaches (Hyperband, ASHA, and random search) for low-budget settings. Most notably, these results establish the promise of \ours, which either outperforms (on \lcbench{} and \taskset{}) or competes closely (on \pdone{}) with \dpl{} and \dyhpo{}. \ours\ is also consistently the best on average rank 
\editmode{across all benchmarks (see Appendix~\ref{app:ablations:pairwise_comp}, Figure~\ref{fig:rebuttal-anytime})}.
Appendix~\ref{a:per_task_plot} (Figures~\ref{fig:per-task-lcb1}-\ref{fig:per-task-taskset2}) offers a closer look at the raw error metrics for each algorithm per task. These detailed results collectively confirm the robustness of \ours\ for HPO tasks, showing its ability to compete if not outperform the most competitive baselines.

\subsection{Ablation on Acquisition}
\label{sec:ablate_acquisition}

\begin{figure*}[!htb]
    \centering
    \includegraphics[width=\linewidth]{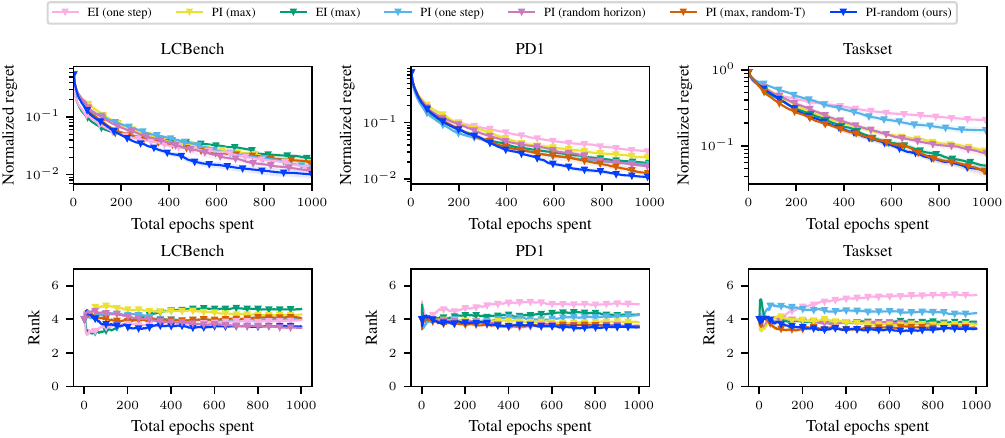}
    \caption{Results of an ablation study of the acquisition function in \ours{} on each benchmark family. First row shows normalized regret aggregated across multiple tasks in each benchmark (Appendix \ref{a:benchmarks}). Second row shows the average ranks of each method.}
    \label{fig:af_ablate}
\end{figure*}

In this section, we evaluate how \ours{} performs in combination with other acquisition functions and aim to assess to what extent our novel acquisition function described in Section~\ref{sec:acquisition} contributes to its HPO success. To this end, we compare against \ours{} variants combining \ftpfn{} with 
the EI-based acquisitions used in prior-art~\citep{wistuba-neurips22,kadra-neurips23}, i.e., \mfei{} predicting one step in the future (\dyhpo{}), and \eimax{} predicting at the highest budget $b_{\text{max}}$ (\dpl{}). We also include their PI counterparts \mfpi{} and \pimax{}, as well as variants of \mfpirand{} that only vary the prediction horizon, \pirandh{} with $T=0$, or only vary the threshold, \pirandt{} with $h = b_{\text{max}}$. Apart from the methods compared, the experimental setup is identical to that in Section~\ref{sec:sota}.

\paragraph{Results discussion:} Figure~\ref{fig:af_ablate} shows the comprehensive results for our $\clftbo{}$ variants for each of the benchmarks, in terms of average ranks and average normalized regrets, aggregated across all tasks. Generally, we find that performance varies strongly between acquisitions, suggesting this choice is at least as important for HPO success as our surrogate's superior predictive quality. In particular, we find that combinations with the EI-based acquisitions \mfei{} and \eimax{} proposed in prior-art, are amongst the worst-performing variants, both in terms of rank and regret. For example, \eimax{} fails on \lcbench{} and \pdone{}, while \mfei{} fails on \pdone{} and \taskset{}. Curiously, these trends do not seem to extend to our baselines, e.g., \dpl{} using \eimax{} performs strongly on \lcbench{} and \dyhpo{} using \mfei{} performs strongly on \pdone{}. We conjecture that this failure is related to the (justified) lack of confidence $\ftpfn{}$ has about its predictions, as is evident by the superior log-scores in Table~\ref{tab:quality_predictions}. As a result, the predicted posterior will be heavy-tailed, resulting in the high EI values for those configurations our predictions are least confident for. While this drives exploration, in the very low budget regime, it can easily lead to a catastrophic failure to exploit. Overall, we find that while some variants are successful at specific tasks in early stages of the optimization, none exhibit the same robustness in performance across benchmarks, making \mfpirand{} the clear winner.

\editmode{We perform comparable ablation studies for \dpl{} and \dyhpo{}, as detailed in Appendix~\ref{app:ablations:acq_baseline}, to demonstrate the benefits of randomizing the horizon and the threshold.}

\section{Conclusion}
\label{sec:conclusion}
In this paper, we proposed \ftpfn{} a novel surrogate for freeze-thaw Bayesian optimization. We showed that the point and uncertainty estimates produced by \ftpfn{} through \emph{in-context learning} are superior to those obtained by fitting/training recently proposed deep Gaussian process~\citep{wistuba-neurips22} and deep power law ensemble~\citep{kadra-neurips23} models, while being over an order of magnitude faster. We presented the first-ever empirical comparison of different freeze-thaw implementations. Our results confirm the superiority of these HPO methods, in the low-budget regime, and show that our in-context learning approach is competitive with the state-of-the-art.

Despite our promising results,
we admit that attaining the sample efficiency required to scale up to modern deep learning (e.g., LLM pretraining) remains a challenging endeavor. Future work should attempt to extend our approach to take advantage of additional sources of prior information, e.g., 
to do in-context meta-learning, 
leveraging learning curves on related tasks~\citep{ruhkopf-tmlr22}; 
to incorporate user priors~\citep{muller-icml23a,mallik2023priorband};
and additional information about the training process (e.g., gradient statistics).
An alternative to scaling up is scaling in parallel. Here, we expect our in-context learning approach to reduce the overhead further, as the online learning/refitting stage occurs on the critical path, while in-context learning during prediction can easily be parallelized.
Finally, the current \ftpfn{} model has some limitations. First, 
it requires the performance metric and each hyperparameter value to be normalized in $[0,1]$; and supports up to 10 hyperparameters. While we believe that this is reasonable, future work building systems should push these limits, train larger models, on more data, and explore ways to scale to larger context sizes.
In summary, there is a lot left unexplored, and we hope that the relative simplicity, efficiency, and public availability of our method, lowers the threshold for future research in this direction.

\section*{Impact Statement}
This paper presents work whose goal is to advance the field of hyperparameter optimization (HPO) in machine learning. There are many potential societal consequences of machine learning, none which we feel must be specifically highlighted here. We would, however, like to highlight that our work makes HPO more robust and efficient, and will thus help make machine learning more reliable and sustainable.

\section*{Acknowledgements}
We thank Johannes Hog for his feedback on an earlier version of the paper.
Frank Hutter is a Hector Endowed Fellow at the ELLIS Institute T\"{u}bingen.
All authors acknowledge funding by 
the state of Baden-W\"{u}rttemberg through bwHPC, the German Research Foundation (DFG) through grant numbers INST 39/963-1 FUGG and 417962828, and
the European Union (via ERC Consolidator Grant Deep Learning 2.0, grant no.~101045765), TAILOR, a project funded by EU Horizon 2020 research and innovation programme under GA No 952215. Views and opinions expressed are however those of the author(s) only and do not necessarily reflect those of the European Union or the European Research Council. Neither the European Union nor the granting authority can be held responsible for them.

\begin{center}\includegraphics[width=0.3\textwidth]{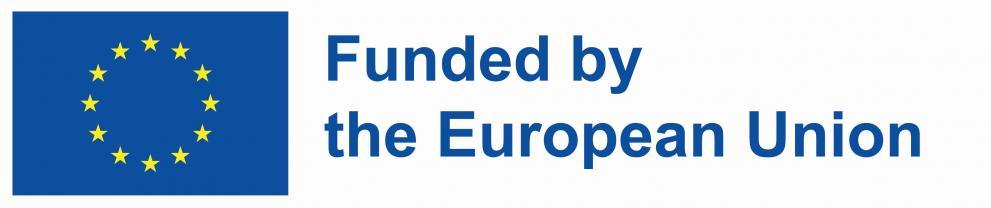}\end{center}

\newpage
\bibliographystyle{icml2024/icml2024}
\bibliography{icml2023/bib/lib,icml2023/bib/local,icml2023/bib/proc,icml2023/bib/strings}

\newpage
\appendix
\onecolumn

\section{Further Details about \ours{}}
\label{a:ftpfn}

\subsection{The Learning Curve Prior}
\label{a:lcprior}
In this section, we continue the discussion of our learning curve prior defined by Equation~\ref{eq:priorcurve}:
\begin{align*}
&\priorcurve(\lambda, t) \sim \mathcal{N}\left( f_{\text{comb}}(t; \mathcal{E}), \sigma^2 \right) \notag 
&\text{with} \quad f_{\text{comb}}(t; \mathcal{E}) = y_0 + (y_{\infty} - y_0) \cdot \sum_{k=1}^K w_k f_k(t; \Psi_k) \notag \\
&&\text{and} \quad (\sigma^2, \underbrace{\left( y_{\infty}, w_1, \ldots, w_K, \Psi_1, \ldots, \Psi_K \right)}_{\mathcal{E}}) \sim \priorconfig(\lambda; \theta)
\end{align*}
Note that $\sigma^2$ and $\mathcal{E}$ are all outputs of the same neural network $\priorconfig$. Due to the symmetry of this network, when marginalizing over $\lambda$ and $\theta$, all these parameters would have the same distribution. This is undesirable. To impose parameter-specific marginal distributions, we (i) estimate the empirical CDF of marginal output distribution; (ii) apply it to each output to obtain a new output with $\mathcal{U}(0,1)$ marginal distribution; (iii) apply the icdf of the parameter-specific target marginal distribution. Specifically, let $u_1,u_2,u_3 \sim \mathcal{U}(0,1)$ be three i.i.d. uniform random variables that are hyperparameter independent and like $\theta$ are sampled once per task, then the non-basis curve specific parameters of our curve model are (marginally) distributed as follows: 
\begin{align*}
y_{\infty} \sim \mathcal{U}(y_0,y_{max}) \quad \text{with} \quad
y_0 = \min(u_1,u_2) \quad \text{and} \quad
y_{max} = \begin{cases}
        \max(u_1,u_2) & \text{if } u_3 \leq 0.25\\
        1.0 & \text{if } u_3 > 0.25
    \end{cases} \\
    \log(\sigma) \sim \mathcal{N}(-5,1) \quad \quad \quad \quad
    w_k = \frac{W_k}{W} \quad \text{with} \quad W_k \sim \mathrm{Gamma(1,1)} \quad \text{and} \quad W = \sum_{k=1}^K W_k
\end{align*}
Each of the basis curves takes the form
\begin{equation*}
f_k(t; \Psi_k) = f_k'(x_t;\Psi_k)
\text{\quad with \quad }
x_t = 
\begin{cases}
        t & \text{if } t \leq x_{\text{sat},k}^\lambda\\
        r_{\text{sat},k}^\lambda (t - x_{\text{sat},k}) + x_{\text{sat},k} & \text{if } t > x_{\text{sat},k}^\lambda
    \end{cases}
\end{equation*}
where each
$f_k$ has the following four parameters $\Psi_k$:
\begin{description}[labelindent=0.5cm]
\item[$\alpha_{k}$] The skew of the curve, determining the convergence rate change over time.
\item[$x_{\text{sat},k},\, y_{\text{sat},k}$] The point at which model performance saturates and the convergence rate is suddenly reduced.
\item[$r_{\text{sat},k}$] The reduced convergence rate after saturation, which can be negative, modeling divergence.
\end{description}
and $f_k'(x_t,\theta_k)$ is a [0,1] bounded monotonic growth function. The formulas for these growth functions, alongside the target distributions of their parameters, are listed in Table~\ref{table:basiscurves}.

\begin{figure*}
  \includegraphics[width=\textwidth]{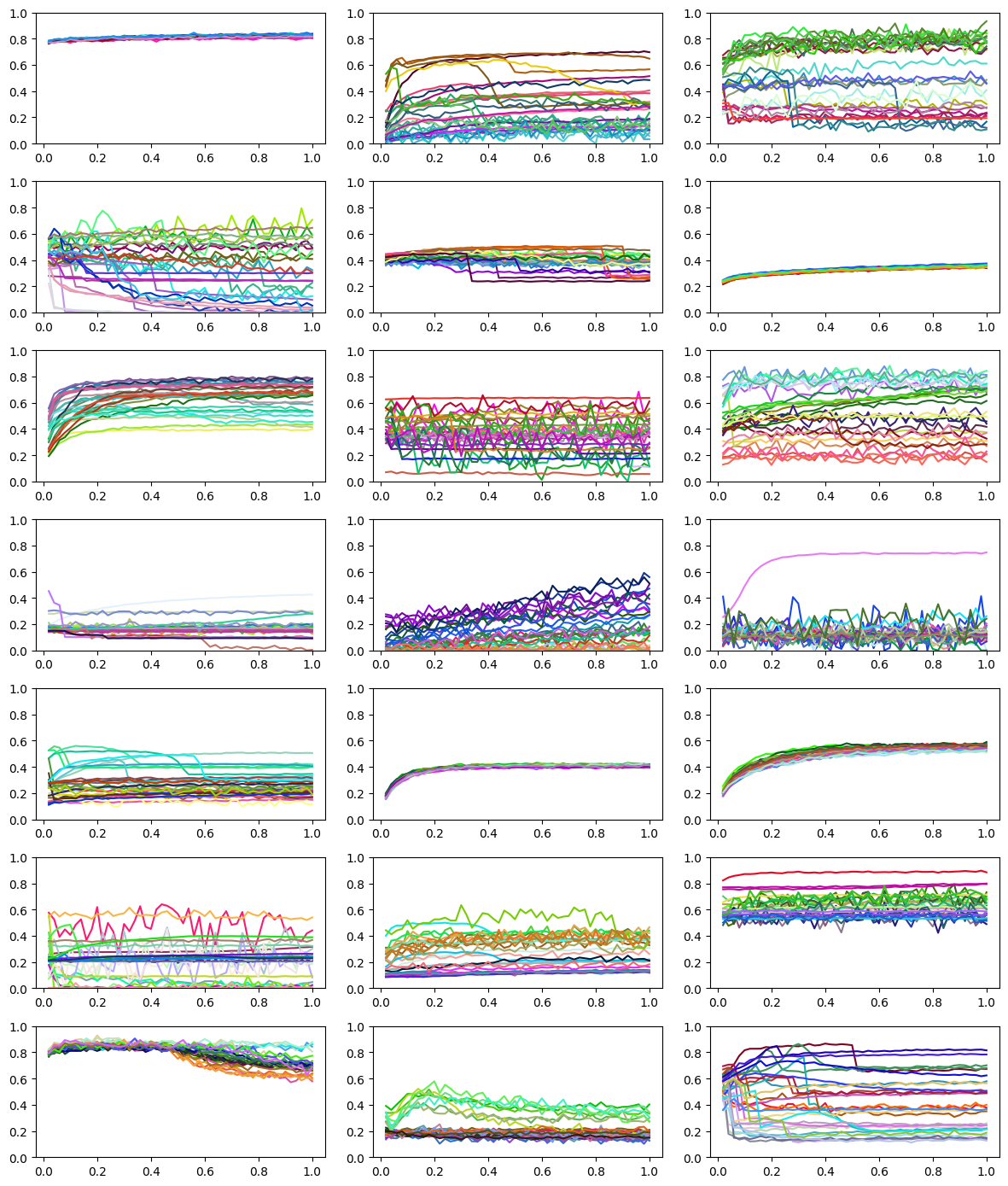}
  \caption{Twenty-one i.i.d. samples of the \ftpfn{} prior, i.e., synthetically generated collections of learning curves for the same task using different hyperparameter configurations. In these examples, we consider 3 hyperparameters that are mapped onto the color of the curves, such that runs using similar hyperparameters, have similarly colored curves. We observe correlations, in varying degrees, between curves on the same task, especially with similar hyperparameter configurations.}
  \label{fig:prior_sample}
\end{figure*}

Finally, note that given these choices, we have $f_{\text{comb}}(t, \mathcal{E}) \in [0,1]$ and we clip the Gaussian noise in $\priorcurve(\lambda,t)$ in the same range. As a consequence, if performance does not naturally fall in this range, it must be normalized before passing it to \ftpfn{}. Examples of collections of curves generated using this prior can be found in Figure~\ref{fig:prior_sample}.

\begin{table}[hbp]
\centering
\begin{tabular}{lll | l} 
\hline
\textbf{Reference name} & \textbf{Formula} $f'_k(x_t;\Psi_k)$ & \multicolumn{2}{l}{\textbf{Prior} $p(\Psi_k)$}  \\ 
\hline
$\text{pow}\textsubscript{4}$& $1-((\epsilon_1^{\frac{-1}{\alpha_1}}-1)*\frac{x_t}{x_{\text{sat},1}} + 1)^{-\alpha_1}$      & $\ln(\alpha_1) \sim \mathcal{N}(1,1)$ & $\log_{10}(x_{\text{sat},k}) \sim \mathcal{N}(0,1) ,\, \forall \, k$                       \\
$\text{exp}\textsubscript{4}$  &   $1-(\epsilon_2)^{(\frac{x_t}{X_{\text{sat},2}})^{\alpha_2}} $  & $\ln(\alpha_2) \sim \mathcal{N}(0,1)$ & $\log_10(\epsilon) \sim \mathcal{U}(-3,0),\, \forall \, k$                          \\
$\text{ilog}\textsubscript{4}$   &  $1-\frac{\ln(\alpha_3)}{\ln((\alpha_3^{\frac{1}{\epsilon_3}} - \alpha_3) \frac{x_t}{X_{\text{sat},3}} + \alpha_3)}$   & $\ln(\alpha_3 - 1) \sim \mathcal{N}(-4,1)$ &  $y_{\text{sat},k} = y_{\infty} - \epsilon \cdot (y_{\infty} - y_0),\, \forall \, k$                         \\
$\text{hill}\textsubscript{4}$ &  $1-\frac{1}{(\frac{x_t}{X_{\text{sat},4}})^{\alpha_4}(\frac{1}{\epsilon_4}-1) + 1}$   & $\ln(\alpha_4) \sim \mathcal{N}(0.5,0.25)$ & $1 - r_{\text{sat},k} \sim Exp(1),\, \forall \, k$           \\
\hline
\end{tabular}
\caption{The formulas for each of the four basis functions in our curve prior. Note that each of them are normalized to start at 0, converge to 1, and pass through the saturation point.}
\label{table:basiscurves}
\end{table}

\subsection{Meta-training Data Generating Procedure}
\label{a:priorgen}
A single meta-training example in our setting corresponds to a training set $D_{\text{train}}$ and test set $D_{\text{test}}$, where ${D_{\text{train}}=\bigcup_{\lambda \in \Lambda}\bigl\{\bigl((\lambda, \frac{b}{\bmax}), \priorcurve(\lambda, \frac{b}{\bmax})\bigl)\bigl\}_{b=1}^{b_\lambda}}$ corresponds to the (synthetic) partial learning curves observed thus far (i.e., the analog of $H$ at test time) and ${D_{\text{test}} \subseteq \bigcup_{\lambda \in \Lambda}\{((\lambda, \frac{b}{\bmax}), \priorcurve(\lambda, \frac{b}{\bmax}))\}_{b=b_\lambda}^{\bmax}}$ the extrapolation targets we want \ftpfn{} to predict. To keep the input size of \ftpfn{} fixed we choose $|D_{\text{train}}| + |D_{\text{test}}| = N = 1,000$ and vary the size of $|D_{\text{train}}| \sim \mathcal{U}(0,N-1)$. As $\bmax$ varies in practice, we sample it log-uniformly in $[1,N]$. Note that in the special case $b_{\text{max}} = 1$, we train \ftpfn{} for black box BO. $\Lambda = \{\lambda_i\}_{i=1}^N$ is our synthetic configuration space with $\lambda_i \sim \mathcal{U}(0,1)^m$, with $|\lambda_i| = m \sim \mathcal{U}(0,M)$ the dimensionality of our configuration space. We determine $b_\lambda$ by sampling a bag of $|D_{\text{train}}|$ elements from $\Lambda$ proportionally to weights $\{w_\lambda\}_{\lambda \in \Lambda}$ that follow a Dirichlet distribution with $\log_{10}(\alpha) \sim \mathcal{U}(-4,-1)$ resulting in heterogeneous budget allocations that vary from breadth-first to depth-first.\footnote{We adopt the same strategy to generate benchmark tasks for our evaluation of prediction quality described in Section~\ref{sec:quality_surrogate}.} We use the same weights to sample another bag of $|D_{\text{test}}|$ determining the number of extrapolation targets for each $\lambda$, where each target $b$ is chosen $\mathcal{U}(b_\lambda,b_{\text{max}})$. Finally, to generate the corresponding performance observation/target, we first instantiate the random variables that are task-specific but do not depend on $\lambda$, i.e., $y_0$, $y_{\text{max}}$ and the architecture and weights $\theta$ of the neural network $\priorconfig$; and subsequently obtain $\priorcurve(\lambda, \frac{b}{\bmax})$ using Equation~\ref{eq:priorcurve}.

\paragraph{Limitations:} With these modeling choices come some limitations. First, \ftpfn{} is trained for HPO budgets $B \leq N = 1,000$; requires the performance metric $f$ and each hyperparameter value to be normalized in [0,1]; and supports up to $M=10$ hyperparameters.

\subsection{Architecture and Hyperparameters}
\label{a:pfntraining}
Following \citet{muller-iclr22a}, we use a sequence Transformer~\citep{vaswani-neurips17a} for \ftpfn{} and treat each tuple ($\lambda$, $t$, $\priorcurve$($\lambda$, $t$)) (for train) and ($\lambda$, $t$) (for test) as a separate position/token. We do not use positional encoding such that we are permutation
invariant. \ftpfn{} outputs a discretized approximation of the PPD, each output corresponding to the probability density of one of the equal-sized bins. We set the number of bins/outputs to 1,000. For the transformer, we use 6 layers, an embedding size of 512, four heads, and a hidden size of 1,024, resulting in a total of 14.69M parameters. We use a standard training procedure for all
experiments, minimizing the cross-entropy loss from Equation~\ref{eq:celoss} on 2.0M  synthethic datasets generated as described in Section~\ref{a:priorgen}, using the Adam optimizer~\citep{kingma-nips15a} (learning rate 0.0001, batch
size 25) with cosine annealing~\citep{loshchilov-iclr17a} with a linear warmup over the first
25\% epochs of the training. Training took roughly 8 GPU hours on an RTX2080 GPU and the same \ftpfn{} is used in all experiments described in Section~\ref{sec:exp}, without any retraining/fine-tuning.

\vspace{3cm}
\subsection{Acquisition function}\label{app:af}

\begin{figure}
\centering
\includegraphics[width=0.8\textwidth]{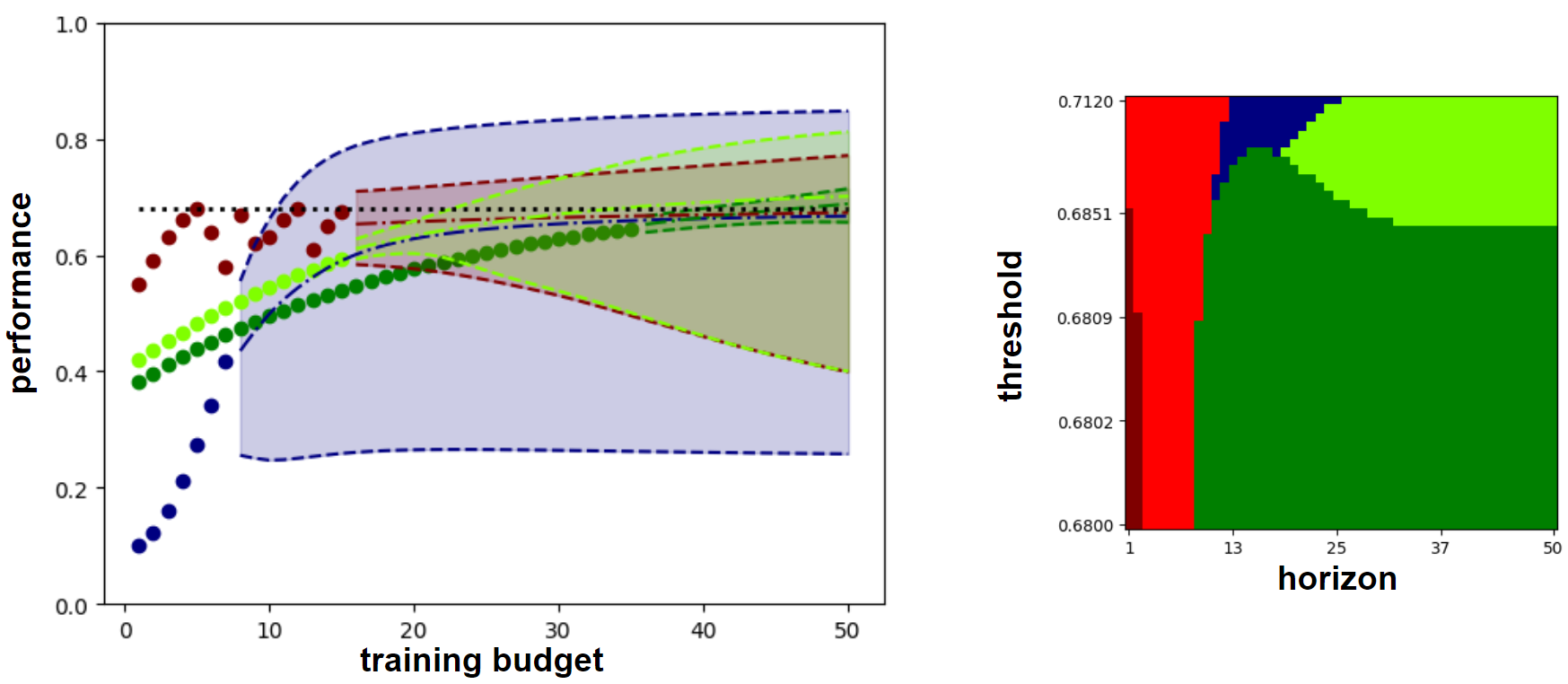}
 \caption{Illustration of the $\texttt{MFPI}$ acquisition (Equation~\ref{eq:mfpi}). 
 (\textit{Left}) The figure shows a collection of partial learning curves and their corresponding continuations predicted by our \ftpfn{} model. Here again, we consider 3 hyperparameters whose values are mapped onto the color of the curves. 
 (\textit{Right})The figure shows the color of the curve continued (i.e., maximizing $\texttt{MFPI}$) for different values of the horizon and threshold parameters. Note that the ranges shown (and scale used), match those sampled uniformly by $\texttt{MFPI-random}$ (Equation~\ref{eq:mfpi_random}) and consequently, the likelihood of continuing a specific curve is proportional to the surface area covered in this image by its corresponding color. Finally, note that the bright red color corresponds to starting a new curve.}
 \label{fig:mfpi_example}
\end{figure}

Algorithm~\ref{algo:mf-af} describes the acquisition procedure \mfpirand{}, used in \clftbo. 
In each iteration of \clftbo{} (L\ref{algo:ftbo:startloop}-L\ref{algo:ftbo:endloop} in Algorithm~\ref{algo:ftbo}), Algorithm~\ref{algo:mf-af} is invoked once taking as input the configuration space $\Lambda$, the surrogate model $\surrmodel$, the observed history $H$, and the maximal training steps $\bmax$ of a configuration. 
First, the random horizon $h^{\text{rand}}$ and the scaled factor of improvement $\tau^{\text{rand}}$ (and thereby $T^{\text{rand}}$) are sampled once in every execution of the algorithm (L\ref{algo:mf-af:samplehorizon}-L\ref{algo:mf-af:scalethreshold}). 
This process can be seen as instantiating an acquisition function from a portfolio of multi-fidelity PIs.
The choice of PI, the multi-fidelity component of extrapolating hyperparameters, and the random selection of an acquisition behaviour lends the naming of this acquisition function, \mfpirand.
Then, for each candidate hyperparameter $\lambda \in \Lambda$, the performance of the hyperparameter at a total step of $b_{\lambda}+h^{\text{rand}}$ is inferred, using the surrogate~$\surrmodel{}$. 
Finally, the candidate with the highest obtained PI score is returned as the candidate solution to query next in the main Algorithm~\ref{algo:ftbo} loop. Figure~\ref{fig:mfpi_example} illustrates the behavior of \mfpirand{} \textit{w.r.t} some values of $h^{\text{rand}}$ and $T^{\text{rand}}$, with \ftpfn{} as a surrogate.

  \begin{algorithm}[H]
  \small
  \caption{\editmode{\mfpirand{}} \label{algo:mf-af}}
  \textbf{Input:} 
    configuration space $\Lambda$, \\
    \hspace*{0.95cm}probabilistic surrogate $\surrmodel$, \\
    \hspace*{0.95cm}history of observations $H$, \\
    \hspace*{0.95cm}maximal steps $\bmax$ \\
    
   \textbf{Output:} $\lambda \in \Lambda$, hyperparameter to evaluate next \\
  
  \textbf{Procedure} \mfpirand($\Lambda$, $\editmode{\surrmodel}$, $H$, $b_\text{max}$): \\
  \begin{algorithmic}[1]

      \STATE $f_{\text{best}} \leftarrow \max \, \{y\}_{(\cdot,\cdot,y) \in H}$ \hfill{\color{gray} best score seen in $H$}

      \STATE $h^{\text{rand}} \sim \mathcal{U}(1, \bmax)$\hfill{\color{gray}random horizon}\label{algo:mf-af:samplehorizon}


      \STATE $T^{\text{rand}} = f_{\text{best}} + 10^{\tau^{\text{rand}}} \cdot (1 - f_{\text{best}}) \quad\text{with}\quad\tau^{\text{rand}} \sim \mathcal{U}(-4, -1)$\hfill{\color{gray}random threshold scaling}\label{algo:mf-af:scalethreshold}
    \STATE \textbf{return} $\underset{\lambda\in\Lambda}{\argmax}\, \mathds{P}(\surrmodel{}(\lambda,~\min(b_{\lambda} + h^{\text{rand}}, \bmax){\color{blue};H}) > T^{\text{rand}})$ \hfill{\color{blue} to perform in-context learning we pass $H$ as input to \ftpfn{}} \\
  \end{algorithmic}
  \end{algorithm}

\section{Benchmarks}
\label{a:benchmarks}
Below, we enumerate the set of benchmarks we have considered. These benchmark cover a variety of optimization
scenarios, including the model being optimized, the task for which it's being trained on, and the training metric with which to optimize hyperparameters with respect to.
Notably, each of these benchmarks are tabular, meaning that the set of possible configurations to
sample from is finite.

This choice of benchmarks is largely dictated by following the existing benchmarks used in prior work, especially the two primary baselines with which we compare to, \dyhpo\ and \dpl. 
These benchmarks were provided using \texttt{mf-prior-bench}\footnote{https://github.com/automl/mf-prior-bench}.

\begin{itemize}
    \item \textbf{\lcbench } \cite{zimmer-tpami21a} [\textit{\dyhpo, \dpl}] - We use all 35 tasks available which
    represent the 7 integer and float hyperparameters of deep learning models from AutoPyTorch.
    Each task represents the 1000 possible configurations, trained for 52 epochs on a dataset taken from the AutoML Benchmark \cite{gijsbers-automl19a}.
    We drop the first epoch as suggested by the original authors.

    \begin{table}[!ht]
    \caption{The 7 hyperparameters for all \lcbench tasks.}
    \begin{center}
    \begin{tabular}{l l l l}
        \toprule
        \textbf{name} & \textbf{type} & \textbf{values} & \textbf{info} \\
        \midrule
            batch\_size & integer & $[16, 512]$ & log \\
            learning\_rate & continuous & $[0.0001, 0.1]$ & log \\
            max\_dropout & continuous & $[0.0, 1.0]$ &  \\
            max\_units & integer & $[64, 1024]$ & log \\
            momentum & continuous & $[0.1, 0.99]$ &  \\
            num\_layers & integer & $[1, 5]$ &  \\
            weight\_decay & continuous & $[1e\mbox{-}05, 0.1]$ &  \\
        \bottomrule
    \end{tabular}
    \end{center}
    \end{table}

    \item \textbf{\taskset } \cite{metz-arxiv20b} [\textit{\dyhpo, \dpl}] This set benchmark provides
    1000 diverse task on a variety of deep learning models on a variety of datasets and tasks.
    We choose the same 12 tasks as used in the \dyhpo\ experimentation which consists of NLP tasks with purely numerical hyperparameters, mostly existing on a log scale.
    We additionally choose a 4 hyperparameter variant and an 8 hyperparameter variant, where the 4 hyperparameter variant is a super set of the former.
    This results in 24 total tasks that we use for the \taskset\ benchmark.

    One exception that needs to be considred with this set of benchmarks is that the optimizers must optimize for is the model's log-loss.
    This metric has no upper bound, which contrasts to all other benchmarks, where the bounds of the metric are known a-priori.
    We note that in the \dyhpo\ evaluation setup, they removed diverging curves as a benchmark preprocessing step, essentially side-stepping the issue that the response function for a given configuration mays return nans or out-of-distribution values.
    As our method requires bounded metrics, we make the assumption that a practitioner can provide a reasonable upper bound for the log loss that will be observed.
    By clampling to this upper bound, this effectively shrinks the range of values that our method will observe.
    As we are in a simulated benchmark setup, we must simulate this a-priori knowledge.
    We take the median value of at epoch 0, corresponding to the median log loss of randomly initialized configurations that have not yet taken a gradient step.
    Any observed value that is nan or greater will then be clamped to this upper bound before being fed to the optimizer.
 
    \begin{table}[!ht]
    \caption{The 4 hyperparameter search space for \taskset.}
    \begin{center}
    \begin{tabular}{l l l l}
        \toprule
        \textbf{name} & \textbf{type} & \textbf{values} & \textbf{info} \\
        \midrule
            beta1 & continuous & $[0.0001, 1.0]$ & log \\
            beta2 & continuous & $[0.001, 1.0]$ & log \\
            epsilon & continuous & $[1e\mbox{-}12, 1000.0]$ & log \\
            learning\_rate & continuous & $[1e\mbox{-}09, 10.0]$ & log \\
        \bottomrule
    \end{tabular}

    \end{center}
    \end{table} 
    \begin{table}[!ht]
    \caption{The 8 hyperparameter search space for \taskset.}
    \begin{center}
    \begin{tabular}{l l l l}
        \toprule
        \textbf{name} & \textbf{type} & \textbf{values} & \textbf{info} \\
        \midrule
            beta1 & continuous & $[0.0001, 1.0]$ & log \\
            beta2 & continuous & $[0.001, 1.0]$ & log \\
            epsilon & continuous & $[1e\mbox{-}12, 1000.0]$ & log \\
            learning\_rate & continuous & $[1e\mbox{-}09, 10.0]$ & log \\
            exponential\_decay & continuous & $[9e\mbox{-}07, 0.0001]$ & log \\
            l1 & continuous & $[1e\mbox{-}09, 10.0]$ & log \\
            l2 & continuous & $[1e\mbox{-}09, 10.0]$ & log \\
            linear\_decay & continuous & $[1e\mbox{-}08, 0.0001]$ & log \\
        \bottomrule
    \end{tabular}

    \end{center}
    \end{table}
    
    \item \textbf{\pdone } \cite{wang-arxiv21a} [\textit{\dpl}] These benchmarks were obtained from the
    output generated by HyperBO ~\citep{wang-arxiv21a} using the dataset and training setup of~\cite{gilmer-github2021}.
    We choose a variety of tasks including the tuning of large vision ResNet~\citep{zagoruyko-bmvc16a}
    models on datasets such as CIFAR-10,
    CIFAR-100~\citep{krizhevsky-tech09a} and SVHN~\citep{liao-arxiv15} image classification datasets, along
    with training a ResNet~\citep{he-cvpr16a} on the ImageNet~\citep{russakovsky-ijcv15a} image classification dataset.
    We also include some natural language processing tasks, notable transformers train on
    the LM1B~\citep{chelba-corr13} statistical language modelling dataset, the XFormer~\citep{lefaudeux-github22}
    trained on the WMT15 German-English~\citep{bojar-wmt15} translation dataset and also a transformer trained to sequence prediction for protein modelling on the uniref50 dataset. Lastly, we also include a simple CNN trained on the
    MNIST~\citep{deng-ieee12} and Fashion-MNIST~\citep{xiao-arxiv17a} datasets.

    Notably, all of these benchmarks share the same 4 deep learning hyperparameters given in table \ref{tab:pdonehps}.

    \begin{table}[!ht]
    \caption{The 4 hyperparameters for all \pdone tasks.}
    \begin{center}
    \begin{tabular}{l l l l}                                                                                                       
        \toprule                                                                                                                       
        \textbf{name} & \textbf{type} & \textbf{values} & \textbf{info} \\                                                             
        \midrule                                                                                                                       
            lr\_decay\_factor & continuous & $[0.01, 0.99]$ &  \\                                                                      
            lr\_initial & continuous & $[1e\mbox{-}05, 10.0]$ & log \\                                                                 
            lr\_power & continuous & $[0.1, 2.0]$ &  \\                                                                                
            opt\_momentum & continuous & $[1e\mbox{-}05, 1.0]$ & log \\                                                                
        \bottomrule                                                                                                                    
    \end{tabular} 
    \label{tab:pdonehps}
    \end{center}
    \end{table}
    
    Each benchmark ranges in the size of their learning curves, depending on the task, ranging from 5 to 1414. For each task, there are different variant based on a pair of dataset and batchsize. In total we evaluate our method on the 16 \pdone\ tasks below.

    \begin{itemize}
        \item \textbf{WideResnet} - Tuned on the CIFAR10, CIFAR100 datasets, each with a constant batch size of $256$ and $2048$.
        Also included is the SVHN dataset with a constant batch size $256$ and $1024$.
        \item \textbf{Resnet} - Tuned on ImageNet with three constant batch sizes, $256$, $512$, and $1024$.
        \item \textbf{XFormer} - Tuned with a batch size of $2048$ on the LM1B statistical language modelling dataset.
        \item \textbf{Transfomer Language Modelling} - Tuned on the WMT15 German-English dataset with a batch size of $64$.
        \item \textbf{Transformer Protein Modelling} - Tuned on the uniref50 dataset with a batch size of $128$.
        \item \textbf{Simple CNN} - Tuned on MNIST and Fashion-MNIST with constant batch sizes of $256$ and  $2048$ for each of them.
    \end{itemize}

\end{itemize}
\section{Baselines} \label{app:baselines}

To use \ours{} in practice for an HPO task, please refer to NePS\footnote{\url{https://automl.github.io/neps/latest/}}. All our baselines were developed into the NePS framework that we forked and copied into our setup. 
Below, we describe the basic configuration of these baselines that were included in our experiments. 

All baseline implementations can be found under \texttt{neps} in our experiment code available at: \url{https://github.com/automl/ifBO/tree/icml-2024}.

\subsection{General baselines}\label{app:baselines-general}

We chose random search based algorithms as baselines for the different benchmarks.
This additionally also shows the utility of the different fidelity scheduling algorithms in HyperBand and ASHA which traverses the fidelity space in progressive geometric intervals, relying on strong performance correlation at these fidelity checkpoints.
For these baselines, we chose the existing implementations in NePS, benchmarked in previously published work~\citep{mallik-neurips23a}.

\paragraph{Random Search}

Simply searches uniformly random in the hyperparameter space. 
The fidelity is set to the $b_{\text{max}}$ as specified by each benchmark instance (see, Appendix~\ref{a:benchmarks}).
Therefore, as an example, a budget of $1000$ freeze-thaw steps, will be equivalent to $20$ full random search evaluations for $\lcbench{}$ and $\taskset{}$ tasks.

\paragraph{HyperBand}

The NePS implementation follows the algorithm described in~\citet{li-jmlr18a} and uses the early stopping hyper-hyperparameter, as $\eta=3$.
The $b_{\text{min}}$ is either $1$ or as specified by the benchmark instances. 
Similarly for $b_{\text{max}}$.

\paragraph{ASHA}

The NePS implementation follows the algorithm described in~\citet{li-mlsys20a} and uses the early stopping hyper-hyperparameter, as $\eta=3$.
The $b_{\text{min}}$ is either $1$ or as specified by the benchmark instances. 
Similarly for $b_{\text{max}}$.

\subsection{Freeze-thaw baselines}\label{app:baselines-freeze-thaw}

Here we describe the set of freeze-thaw BO algorithms. We note that due to experimental framework (optimizer-benchmark interfacing and analysis) related differences, performing ablation studies on the original implementations of DyHPO and DPL were not straightforward. 
For consistency and reducing confounding factors, all experiments were performed with implementations in the same experimental framework.
Each of the algorithms were implemented in our custom NePS framework.


\paragraph{Freeze-Thaw with GPs}

This algorithm is designed to take one unit step per configuration in the fidelity space.
The first $3$ samples are selected uniformly random, as influenced by the seed.
Subsequently, a Gaussian Process (GP) is fit on the joint hyperparameter and fidelity space to predict the loss, as a surrogate model.
This baseline uses the greedy MF-EI acquisition function from~\citet{wistuba-neurips22}.
The GP here uses a standard 5/2-Matérn kernel with a lengthscale of $1.0$.

\paragraph{DyHPO}

This implementation follows the exact details given in~\citet{wistuba-neurips22} and their publicly available code\footnote{\url{https://github.com/releaunifreiburg/DyHPO/tree/main}}.
For a quick hyper-hyperparameter glance, refer here: \url{https://github.com/automl/ifBO/blob/icml-2024/src/pfns_hpo/pfns_hpo/configs/algorithm/dyhpo-neps-v2.yaml}

\paragraph{DPL}

This implementation follows the exact details given in~\citet{kadra-neurips23} and their publicly available code\footnote{\url{https://github.com/releaunifreiburg/DPL/tree/main}}.
For a quick hyper-hyperparameter glance, refer here: \url{https://github.com/automl/ifBO/blob/icml-2024/src/pfns_hpo/pfns_hpo/configs/algorithm/dpl-neps-max.yaml}

\section{\editmode{Further Ablations}}\label{app:ablations}

\subsection{\editmode{Effectiveness of modeling curve divergence}}\label{app:ablations:broken}

\begin{figure}[ht]
    \centering
    \includegraphics{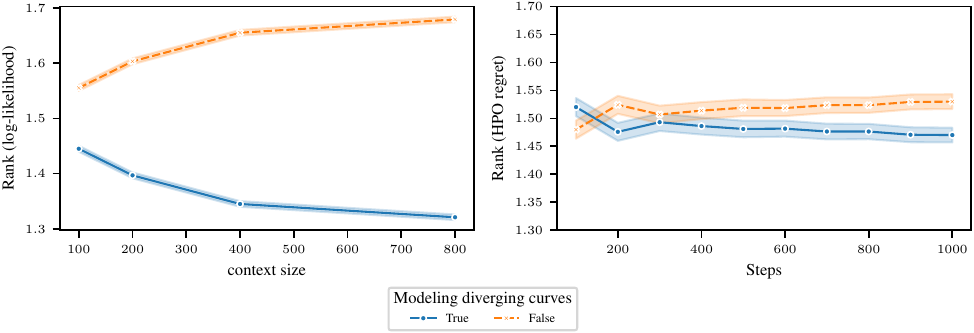}
    \caption{
        \editmode{Comparison of the relative ranks of the performance gained by modeling divergences in ICL-FT-PFN. The plots, showing the average ranks across all the benchmarks (LCBench, PD1, and TaskSet), confirm the merits of capturing diverging curves both in terms of the quality of the predictions (log-likelihood, left) and HPO performances (regret, right).} 
    }
    \label{fig:rebuttal-broken}
\end{figure}

\editmode{As detailed in Section~\ref{a:lcprior}, our curve prior is capable to model learning curve with diverging behavior. This capability is novel compared to the related works~\cite{adriaensen-neurips23,klein-iclr17a,kadra-neurips23,domhan-ijcai15a}, which are restricted to monotonic curves only. In Figure~\ref{fig:rebuttal-broken}, we empirically show that modeling diverging curves yields a better surrogate model in terms of both extrapolation and HPO.}

\subsection{\editmode{Pairwise comparison of freeze-thaw approaches}}\label{app:ablations:pairwise_comp}

\begin{figure}[ht]
    \centering
    \includegraphics[width=\textwidth]{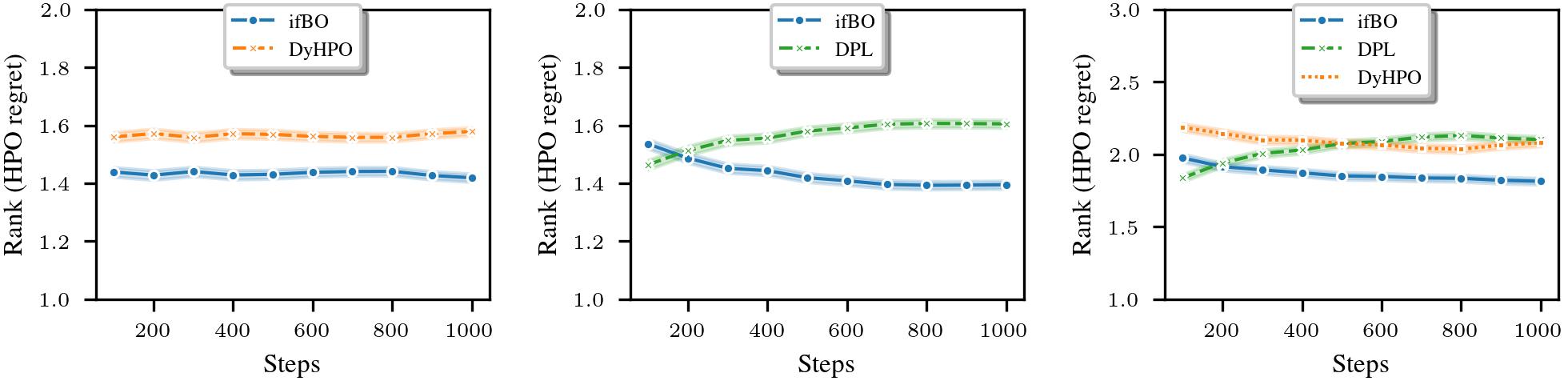}
    \caption{
        \editmode{Comparison of relative ranks when aggregated over \textit{all benchmark families}, showing strong \textit{anytime} performance in both pairwise comparisons and also overall among freeze-thaw algorithms.}
    }
    \label{fig:rebuttal-anytime}
\end{figure}

\editmode{For a fine-grained assessment of the performance of \clftbo{}, we present a pairwise comparison with the main freeze-thaw approaches including \dpl{} and \dyhpo{}. This is to visualize the relative gain of performance compared to each baseline, which may have been hidden from Figure~\ref{sec:sota}. As shown in Figure~\ref{fig:rebuttal-anytime}, our approach dominates consistently \dpl{} and \dyhpo{} after $\approx$ 150 steps of HPO run. 
}

\subsection{\editmode{Acquisition function ablation of the baselines}}\label{app:ablations:acq_baseline}

\begin{figure}[ht]
    \centering
    \includegraphics[width=0.9\textwidth]{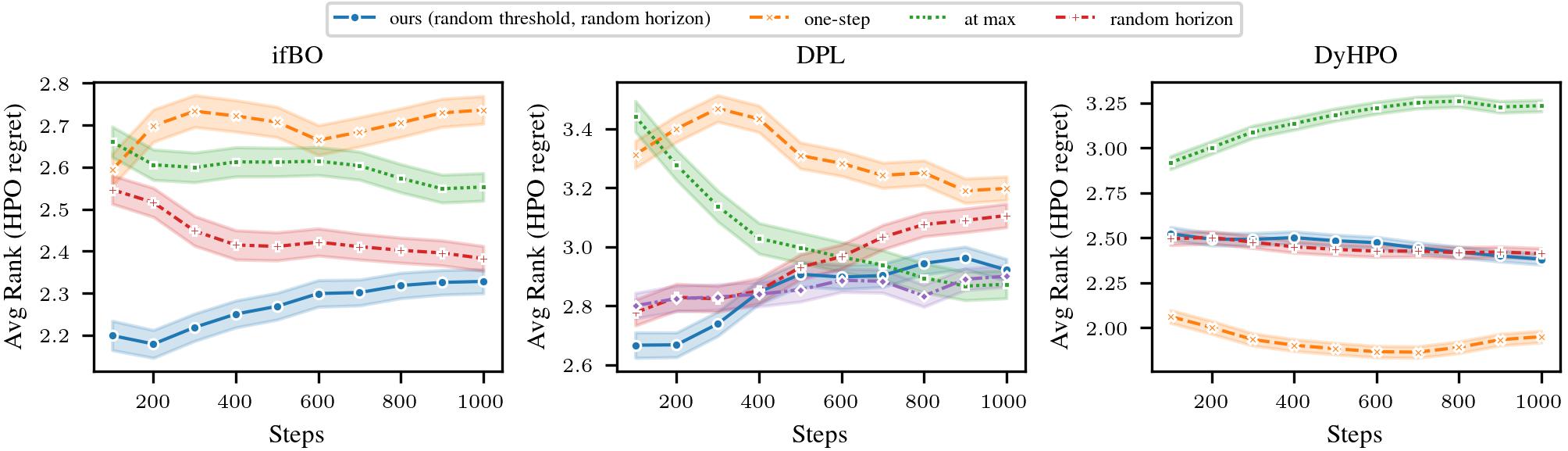}
    \caption{
        \editmode{Relative ablation over the horizon and threshold parameters of a multi-fidelity \texttt{AF}.
        For each algorithm, we take the \texttt{AF} designed into the original algorithm and ablate over the two variables: extrapolation horizon and the best performance threshold.
        }
    }
    \label{fig:rebuttal-af-random}
\end{figure}

\editmode{In this section, our objective is to explore the impact of incorporating randomization into the acquisition function on the baseline methods (\dpl{} and \dyhpo{}). For this purpose, we assess each baseline across four distinct acquisition functions (Figure~\ref{fig:rebuttal-af-random}). The variants include: (\textit{ours}), where both the horizon and the threshold for improvement are randomly selected, similar to the approach in \clftbo{}; (\textit{one-step}), where the horizon and threshold for improvement are chosen as in \dyhpo{}; (\textit{at max}), where the selection criteria for the horizon and threshold follow the methodology in \dpl{}; and (\textit{random horizon}), where the horizon is randomly determined, and the threshold is set to the best value observed.}

\editmode{The results presented in Figure~\ref{fig:rebuttal-af-random} confirm that the randomization technique markedly enhances the performance of methods capable of extending learning curves over many steps, such as \clftbo{} and \dpl{}. Furthermore, please note that only the greedy \textit{one-step} acquisition function is effective for \dyhpo{}, given that it is specifically designed for one-step ahead predictions.}

\subsection{\editmode{Comparison of freeze-thaw approaches with \mfpirand{}}}\label{app:ablations:acq_mfpirandom}

\begin{figure}[ht]
    \centering
    \includegraphics[width=0.9\textwidth]{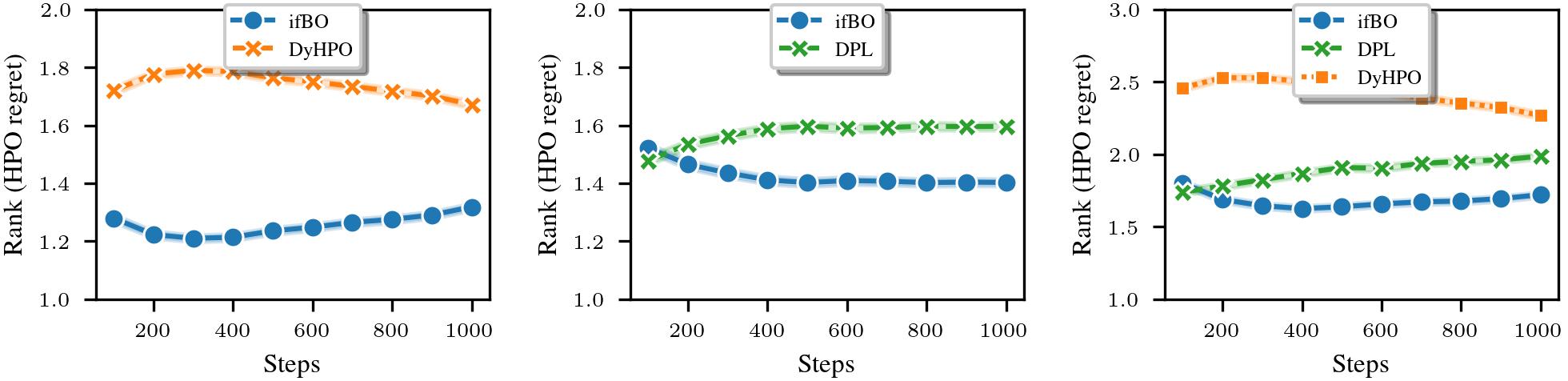}
    \caption{
        \editmode{Comparison of freeze-thaw approaches (\clftbo{}, \dpl{}, and \dyhpo{}) when using our acquisition (\mfpirand{}) with their specific surrogate models. The plots represent the average ranks over all benchmarks (LCBench, PD1, and Taskset). This ablation confirms that our novel surrogate (and not only our novel acquisition function) contributes significantly to the HPO performance of \clftbo{}.}
    }
    \label{fig:rebuttal-anytime-mfpi}
\end{figure}

\editmode{In Figure~\ref{fig:rebuttal-anytime-mfpi}, we present a comparison of freeze-thaw approaches—including ours, \dpl{}, and \dyhpo{}—when employing our acquisition function (\mfpirand{}). Despite all models utilizing the same acquisition function, our model significantly outperforms those of \dpl{} and \dyhpo{}. This clearly indicates the crucial role our prior in achieving the final performance.}

\section{\editmode{Aggregate plots over time}}\label{app:ablations:perf_over_runtime}
\begin{figure}[htbp]
    \centering
    \includegraphics[width=\textwidth]{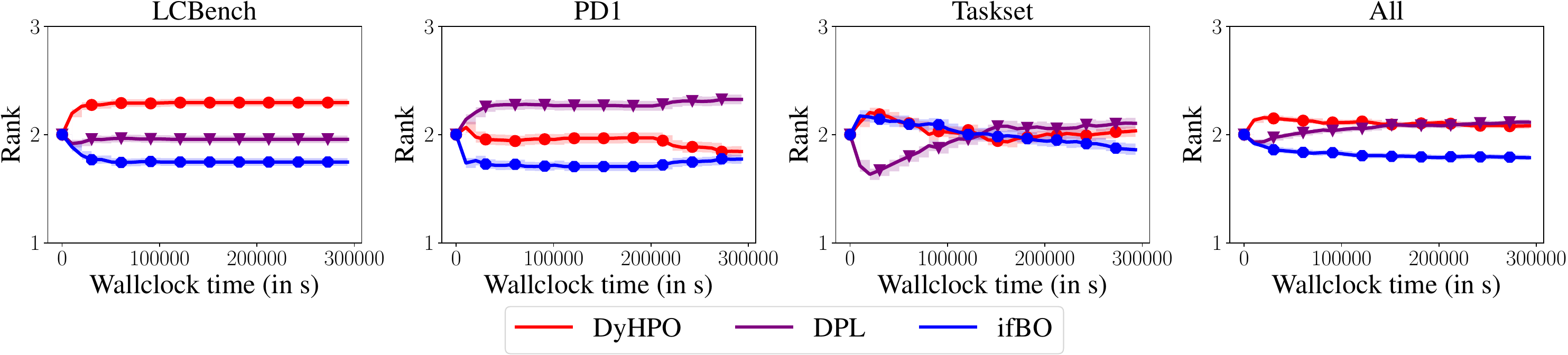}
    \caption{
        Comparing relative rank over wallclock time (in $s$) over different benchmark families and the aggregated result overall. \textcolor{blue}{\clftbo} is on average better than the baselines, \textcolor{red}{\dyhpo} and \textcolor{violet}{\dpl}, except for the TaskSet benchmark family where \textcolor{violet}{\dpl} starts the best but \textcolor{blue}{\clftbo} improves with more budget.
    }
    \label{fig:appendix-wallclock}
\end{figure}

\editmode{
Figure~\ref{fig:appendix-wallclock} plots Figure~\ref{fig:per-bench-sota}(bottom) but with the $x$-axis as cumulative wallclock time from the evaluation costs returned by the benchmark for each hyperparameter for every unit step.
The overall conclusions remain over our HPO budget of $1000$ steps.
\clftbo{} is on average anytime better ranked than the freeze-thaw HPO baselines.
}

\section{Per-task HPO Plots}\label{a:per_task_plot}
\begin{figure*}\centering
    \includegraphics[width=.95\textwidth]{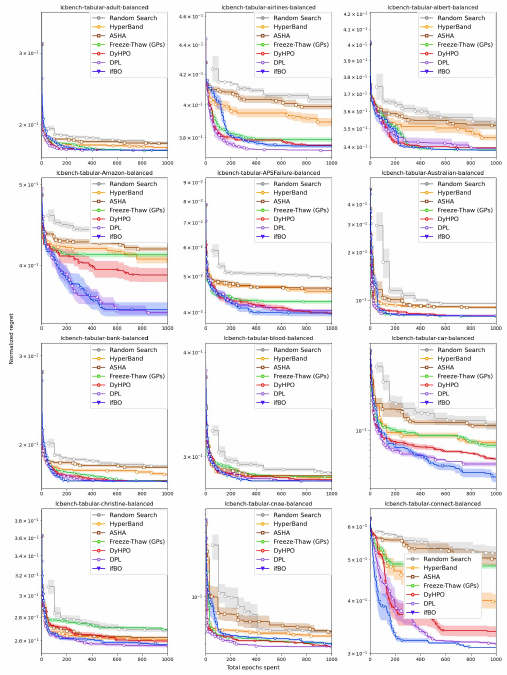}
  \caption{Per-task HPO results on \lcbench{} \label{fig:per-task-lcb1}}
\end{figure*}

\begin{figure*}
\centering
  \includegraphics[width=.95\textwidth]{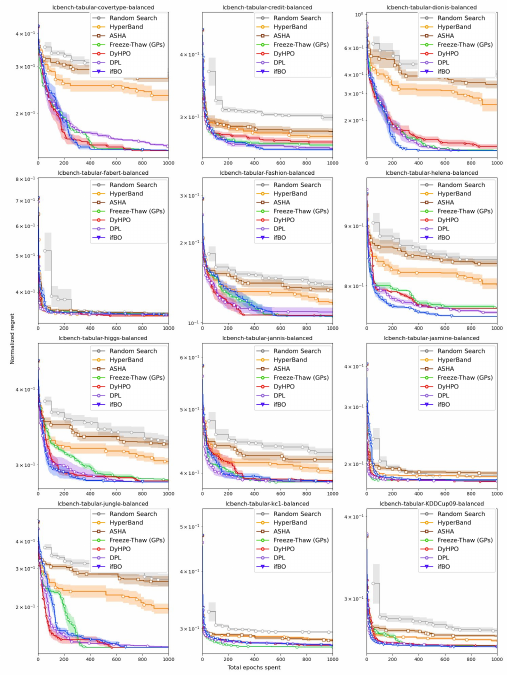}
  \caption{Per-task HPO results on \lcbench{} (cont.)\label{fig:per-task-lcb2}}
\end{figure*}

\begin{figure*}
\centering
 \includegraphics[width=.95\textwidth]{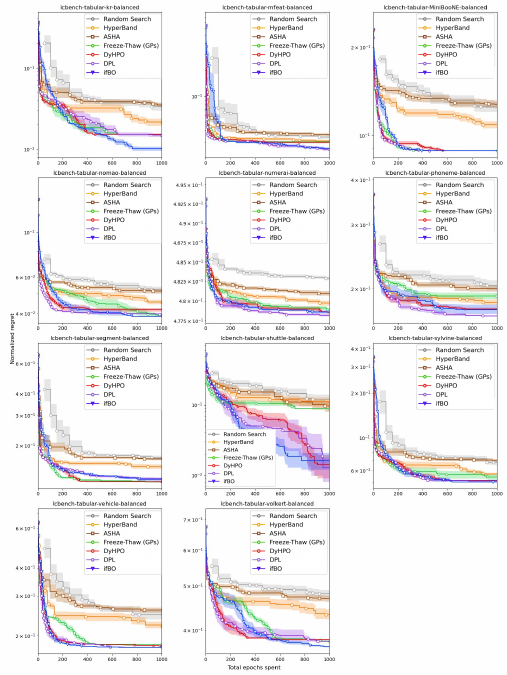}
  \caption{Per-task HPO results on \lcbench{} (cont.)\label{fig:per-task-lcb3}}
\end{figure*}

\begin{figure*}
\centering
\includegraphics[width=.75\textwidth]{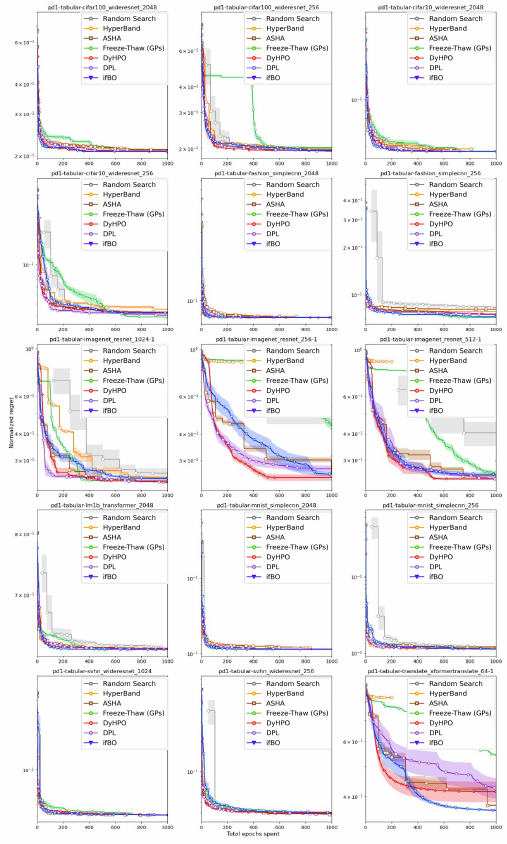}
  \caption{Per-task HPO results on \pdone{} \label{fig:per-task-pd1}}
\end{figure*}

\begin{figure*}
\centering
\includegraphics[width=.95\textwidth]{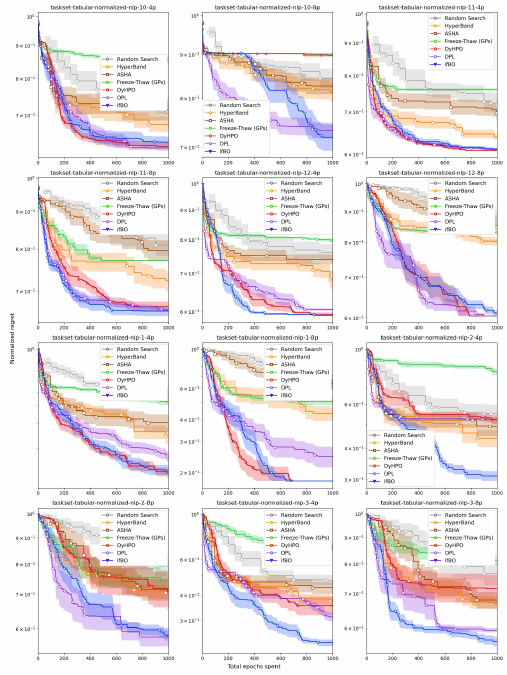}
  \caption{Per-task HPO results on \taskset{} \label{fig:per-task-taskset1}}
\end{figure*}

\begin{figure*}
\centering
\includegraphics[width=.95\textwidth]{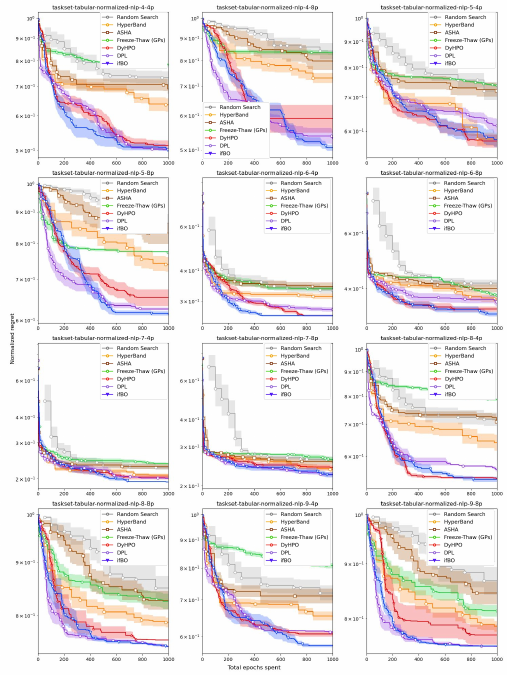}
  \caption{Per-task HPO results on \taskset{} (cont.)\label{fig:per-task-taskset2}}
\end{figure*}

In Section~\ref{sec:hpo}, we presented HPO results on each of these three benchmarks in a comprehensive form, averaging rank and normalized regrets across every task in the suite. These averages may hide / be susceptible to outliers. For completeness, Figures~\ref{fig:per-task-lcb1}-\ref{fig:per-task-taskset2} provide regret plots for every task in the benchmark, averaged across the 10 seeds. We find that our method consistently performs on par, or better than the best previous best HPO method, especially in later stages of the search, without notable outliers.


\end{document}